\documentclass[10pt]{aiaa-tc}
\usepackage{subcaption}
\usepackage{amsfonts}
\usepackage{amsmath}%
\usepackage{amssymb}%
\usepackage{tikz}
\usetikzlibrary{positioning}
\usepackage{morefloats}
\usepackage[section]{placeins}
\usepackage{times}
\newcommand{\pic}[2]{\includegraphics[width=#1\textwidth]{#2}}
 \title{Multiscale Gaussian Processes for Turbulence Modeling}

 \AIAApapernumber{YEAR-NUMBER}
 \AIAAconference{Conference Name, Date, and Location}
 \AIAAcopyright{\AIAAcopyrightD{YEAR}}

 \newcommand{\mb}[1]{\ensuremath{\mathbf{#1}}}
\newcommand{\mc}[1]{\ensuremath{\mathcal{#1}}}
\newcommand{\mr}[1]{\ensuremath{\mathrm{#1}}}

\begin{document}

\maketitle

\begin{abstract}
Standard Gaussian Process (GP) regression, a powerful machine learning tool, is computationally expensive when it is applied to large datasets, and potentially inaccurate when data points are sparsely distributed in a high-dimensional feature space. To address these challenges, a new multiscale, sparsified GP algorithm is formulated, with the goal of application to large scientific computing datasets.  In this approach, the data is partitioned into clusters and the cluster centers are used to define a reduced training set, resulting in an improvement over standard GPs in terms of training and evaluation costs. Further, a hierarchical technique is used to adaptively map the local covariance representation to the underlying sparsity of the feature space, leading to improved prediction accuracy when the data distribution is highly non-uniform.   A theoretical investigation of the computational complexity of the algorithm is presented. The efficacy of this method is then demonstrated on simple analytical functions and on data from a direct numerical simulation of turbulent combustion.
\end{abstract}

\section{Introduction}

The rapid growth in computing power has resulted in the generation of massive amounts of highly-resolved datasets in many fields of science and engineering. 
Against this backdrop, machine learning (ML) is becoming a widely-used tool in the identification of patterns and functional relationships that has resulted in  improved understanding of underlying physical phenomena~\cite{mjolsness2001}, characterization and improvement of models~\cite{yairi1992,ling2015,eric2015}, 
 control of complex systems~\cite{gautier2015}, etc. 
In this work, we are interested in further developing Gaussian Process (GP) regression, which is a popular supervised learning technique.
While GPs have been demonstrated to provide accurate predictions of the mean and variance of functional 
outputs, application to large-scale datasets in high-dimensional feature spaces remains a hurdle.
This is because of the high computational costs and memory requirements during the training stage and 
 the expense of computing the mean and variance during the prediction stage. 
To accurately represent high-dimensional function spaces, a large number of training data points must be used. In this work, an extension to GP regression is developed with the specific goal of applicability in noisy and large-scale scientific computing datasets.

The computational complexity of GPs at the training stage is related to the kernel inversion and the computation of the log-marginal likelihood. For example,
an algorithm based on the Cholesky decomposition is one well-known approach~\cite{chol}. The computational complexity of Cholesky decomposition for a 
problem with $N$ training points is $O\left( N^{3}\right) $. For high-dimensional problems with large $N$, this can be prohibitive.
Even if this cost can be reduced using, for instance, iterative methods~\cite{gumerov-iter}, the 
computation of the predictive mean and variance at $M$ test points will be $O\left(NM\right) $, 
and $O\left(N^2 M\right) $, respectively.
The evaluation (or testing) time could be of great significance if GP evaluations are made frequently during a computational process. 
Such a situation can occur in a scientific computing setting~\cite{transprev,eric2015} where GP evaluations may be performed
every time-step or iteration of the solver.  Reducing both the training and evaluation costs while preserving
the prediction accuracy is the goal of the current work. 

Much effort has been devoted to the construction of algorithms of reduced complexity. Among these is the family of methods of sparse Gaussian process regression. These methods seek to strategically reduce the size of the training set or find appropriate sparse representations of the correlation matrix through the use of induced variables. The covariance matrix, $\mb{\Phi} \in \mathbb{R}^{N\times N}$, contains the pairwise covariances between the $N$ training points. Descriptions for several methods of this family are provided by Qui{\~n}onero-Candela \emph{et al.} \cite{quinonero2005}, who define induced variables $\mb{U}\in \mathbb{R}^{d\times P}$, $\mb{V}\in \mathbb{R}^P$ such that the prediction of a function $f(\mb{q})$ is given by

\begin{equation}
p(f|\mb{U}) \sim \mc{N}(\underline{\phi}_u^T\mb{\Phi}_u^{-1}\mb{V}, 1+\sigma^2-\underline{\phi}_u^T\mb{\Phi}_u^{-1}
\underline{\phi}_u),
\end{equation}

where $\underline{\phi}_u \in \mathbb{R}^P$ is a vector whose elements are $\phi(\mb{u}_i,\mb{q})$, and $\mb{\Phi}_u(m,n) = \phi(\mathbf{u}_m,\mathbf{u}_n)$. $P$ is the number of induced variables used in this representation, and $d$ is the dimensionality of the inputs; i.e. $\mb{u}$, $\mb{q} \in \mathbb{R}^d$. $\mb{q}$ is a single test input vector, and $\mb{u}_m$ is the $m$'th column of $\mb{U}$. $\phi$ is the covariance function. From this, it is evident that if $P<N$, the matrix inversion will be less costly to perform. 

The process of determining the induced variables that can optimally represent the training set can introduce additional computational costs. The most basic method is to randomly select a fraction of data points and define them as the new training set, i.e. choosing $(\mb{U},\mb{V})$ from $(\mb{Q},\mb{y})$, where $\mb{Q} \in \mathbb{R}^{d\times N}$ and $\mb{y}\in \mathbb{R}^{N}$ are the original training inputs and outputs. Seeger \emph{et al.}~\cite{seeger2003}, Smola \emph{et al.}~\cite{smola2001}, and others have developed  selection methods that provide better test results compared to random selection. 
In particular, Walder \emph{et al.}~\cite{walder2008} have extended the concept to include the ability to vary $\phi$ such that its hyperparameters are unique to each $\mb{u}$, i.e. $\phi = \phi_i(\mb{u}_i,\mb{q})$.
These methods often depend on optimizing modified likelihoods based on $(\mb{U},\mb{V})$, or on approximating the true log marginal likelihood \cite{titsias2009}. Methods that introduce new induced variables instead of using a subset of the training points can involve solving optimization problems, again requiring the user to consider the additional computational burden. 

In addition to choosing the inducing variables, one may also change the form of the covariance function from $\phi$ to $\phi'$, where
\begin{equation}
\phi'(\mb{q}_m,\mb{q}_n) = \phi(\mb{q}_m,\mb{U})\mb{\Phi}_u^{-1}\phi(\mb{q}_n,\mb{U})^T.
\end{equation}

In contrast to the case where only a subset of the training data is used, all training points are considered in this approach and the matrices are only of rank $P$. There are many variations on this type of manipulation of the covariance. Alternatively, one can approximate $\mb{\Phi}$ directly by obtaining random samples of the kernel, as is described by Rahimi \emph{et al.}~\cite{recht2007}. However, altering the covariance matrix can cause undesirable behaviors when computing the test variance, since some matrices no longer mathematically correspond to a Gaussian process. The work of Ambikasaran \emph{et al.}~\cite{ambikasaran} improves the speed of GP regression through the hierarchical factorization of the dense covariance matrix into block matrices, resulting in a less costly inversion process.

Other methods of accelerating GP regression focus on dividing the training set into smaller sets of more manageable size. Snelson and Ghahramani \cite{snelson2007local} employ a local GP consisting of points in the neighborhood of the test point in addition to a simplified global estimate to improve test results. In a similar fashion, the work of Park and Choi \cite{park2010} also takes this hierarchical approach, with the global-level GP informing the mean of the local GP. Urtasun and Darrell \cite{urtasun2008} use the local GP directly, without the induced variables. However, this requires the test points to be either assigned to a pre-computed neighborhood of training points~\cite{snelson2007local}, or to be determined on-line~\cite{urtasun2008}. Both of these approaches to defining the neighbourhood can be quite expensive depending on training set size, though massive parallelization may mitigate the cost in some aspects \cite{keane2002} \cite{weiss2014}.

In this work, the data is partitioned into clusters based on modified k-center algorithms and the cluster centers are used to define a reduced training set.  
This leads to an improvement over standard GPs in terms of training and testing costs. This
technique also adapts the covariance function to the sparsity of a given neighborhood of training points. 
The new technique will be referred to as Multiscale GP and abbreviated as MGP.
Similar to the work of Snelson and Ghahramani \cite{snelson2007local}, this flexibility may allow MGP to produce more accurate outputs by means of a 
more informed perspective on the input space. 
The sparse GP technique of Walder \emph{et al.}~\cite{walder2008} can been seen as a very general interpretation of our approach, though a key difference is that the points within the reduced training set originate entirely from the original training set, whereas ~\cite{walder2008} computes new induced variables. Further, in our work, hyperparameters are restricted by the explicit choice and hierarchy of scales in order to streamline the process of their optimization. The work of Zhou \emph{et al.}~\cite{zhou2006} also employs multiple scales of covariance functions, but it does not take into account the neighbourhoods of training points, and neither does it seek to reduce the size of the training set.

The next section of this paper recapitulates the main aspects of standard GP regression. Section 3 introduces the philosophy behind the proposed approach
and presents the algorithm. Section 4 analyses the complexity of the MGP algorithm for both testing and training. 
In section 5, quantitative demonstrations are presented on a simple analytical problems as well as on
 data derived from numerical simulations of turbulent combustion. Following this, key contributions are summarized.

\section{Background on GP regression}
For applications of regression, the dataset consists of sets of points $(\mb{q},y)$, where the input vector $\mb{q} \in\mathbb{R}^{d}$, and the output vector $y\in\mathbb{R}^{c}$. Given a set of $N$ outputs $\mathbf{Y}\in \mathbb{R}^{N\times c}$ and its corresponding set of $N$ inputs $\mathbf{Q}\in \mathbb{R}^{N\times d}$, it is customary\cite{mlbook} to divide them into training and validation sets, $(\mathbf{Q}_{\mr{train}},\mathbf{Y}_{\mr{train}})$ and $(\mathbf{Q}_{\mr{val}},\mathbf{Y}_{\mr{val}})$. These sets consist of $N_{\mr{train}}$ and $N_{\mr{val}}$ points, respectively. A ML algorithm is first applied to the training set to obtain the function $f$, where $f(\mathbf{q})\approx y$. Then, the error on the validation set can be used to adjust any hyperparameters the ML algorithm may possess. Once this is done, the algorithm can be tested by evaluating $f(\mathbf{q}_{\mr{test}})$ and comparing the result with $y_{\mr{test}}$.

In the description that follows, the output $y$ is scalar and assumed to have Gaussian noise with variance $\lambda$.
A GP assumes that the output at the training points, $\mb{Y}_{\mr{train}}$, is drawn from the distribution
\begin{equation}
\mb{Y}_{\mr{train}} \sim \mc{N}(\mb{0},\mb{\Phi} + \lambda\mb{I})
\end{equation}
$\mb{\Phi}$ is a matrix of covariances between training points. The mean of the distribution is unknown, so it is set to zero. The posterior is not limited by any particular prior; setting the mean to zero is for ease of computation only. The noise of the output is added to the covariance between training points because they are assumed to be independent. The current work employs a radial basis function (RBF)\cite{rbfgp1}$^,$\cite{rbfgp2} kernel for computing GP covariance:
\begin{equation}
\mb{\Phi}(m,n) = \phi(\mathbf{q}_m,\mathbf{q}_n) = e^{-\frac{||\mathbf{q}_m - \mathbf{q}_n||^2}{h^2}}
\label{GPeqn}
\end{equation} 
where $\mb{q}_m$, $\mb{q}_n$ are points from $\mb{Q}_{\mr{train}}$. 

Given a test point $\mb{q}_{\mr{test}}$, the corresponding $f(\mb{q}_{\mr{test}})$ can be defined via a conditional probability distribution, $p(f|\mb{Y}_{\mr{train}})$. The mean of this distribution is the predicted value of $f$ given the training points, and the variance is $\sigma^2_f$. 
It is then found that~\cite{GP_book}
\begin{equation}
p(f|\mb{Y}_{\mr{train}}) \sim \mc{N}(\underline{\phi}^T(\mb{\Phi} + \lambda\mb{I})^{-1}\mb{Y}_{\mr{train}}, 1+\lambda-\underline{\phi}^T(\mb{\Phi} + \lambda\mb{I})^{-1}\underline{\phi}),
\end{equation}
where $\underline{\phi}$ is a vector whose elements are $\phi(\mb{q}_{\mr{train},i},\mb{q}_{\mr{test}})$, with $\mb{q}_{\mr{train},i}$ being the $i$th data point in $\mb{Q}_{\mr{train}}$. Given the form of the kernel function, the value of $f$ will be closer to the values of $y_{\mr{train}}$ for which $\mb{q}_{\mr{train}}$ is close to $\mb{q}_{\mr{test}}$.

In cases where $\lambda$ is supplied, $\sigma^2_f$ can be found directly. If there is a different $\lambda$ assigned to each $y_{\mr{train}}$, the scalar $\lambda$ may be replaced by the vector $\mb{\Lambda}$ consisting of all the individual variances at the training points, and the rest of the equations would still hold. In cases where $\lambda$ is unknown, it can be estimated by maximising the log marginal likelihood (LML):
\begin{equation}
\log p(\mb{Y}_{\mr{train}}|\mathbf{Q}_{\mr{train}},h,\lambda) = - \frac{1}{2} \log |\mathbf{\Phi} + \lambda\mb{I}| - \frac{1}{2} \mathbf{Y}^T_{\mr{train}}(\mathbf{\Phi}+\lambda\mb{I})^{-1}\mathbf{Y}_{\mr{train}} -\frac{N_{\mr{train}}}{2}\log(2\pi)
\end{equation}

\textcolor{red}{
Background section
\begin{itemize}
\item Can cut the material a little bit more, perhaps?
\item The last part does not connect to the next section
\end{itemize}
}

\title{Efficient Multiscale Gaussian Process Regression using Hierarchical Clustering
}


\author{Z. Zhang, K. Duraisamy, N. Gumerov
\\
}


\institute{
Department of Aerospace Engineering,\\
University of Michigan, Ann Arbor, MI. \\
              Tel.: +734-615-7270\\
              \email{kdur@umich.edu}           
}

\date{Received: date / Accepted: date}

\maketitle

\begin{abstract}
Standard Gaussian Process (GP) regression, a powerful machine learning tool, is computationally expensive when it is applied to large datasets, and potentially inaccurate when data points are sparsely distributed in a high-dimensional feature space. To address these challenges, a new multiscale, sparsified GP algorithm is formulated, with the goal of application to large scientific computing datasets.  In this approach, the data is partitioned into clusters and the cluster centers are used to define a reduced training set, resulting in an improvement over standard GPs in terms of training and evaluation costs. Further, a hierarchical technique is used to adaptively map the local covariance representation to the underlying sparsity of the feature space, leading to improved prediction accuracy when the data distribution is highly non-uniform.   A theoretical investigation of the computational complexity of the algorithm is presented. The efficacy of this method is then demonstrated on smooth and discontinuous analytical functions and on data from a direct numerical simulation of turbulent combustion.


\end{abstract}

\section{Preliminaries of GP regression}

In this section, a brief review of standard GP regression is presented.
While a detailed review can be found in the excellent book by Rasmussen~\cite%
{rbfgp2}, the description and notations below provide the necessary context
for the development of the multiscale algorithm proposed in this paper.

We consider the finite dimensional weight-space view and the infinite
dimensional function-space view as in Rasmussen~\cite{rbfgp2}. Given a
training set $\mathcal{D}$ of $N$ observations, $\mathcal{D}=\left\{ \left( 
\mathbf{q}_{n},y_{n}\right)~|~n=1,...,N\right\} $, where $\mathbf{q}_{n}\in 
\mathbb{R}^{d}$ is an input vector and $y$ is a scalar output, the goal is
to obtain the predictive mean, $m_{\ast }$, and variance, $v_{\ast }$, at an
arbitrary test point $\mathbf{q}_{\ast }\in \mathbb{R}^{d}$. Training inputs
are collected in the design matrix $Q\in \mathbb{R}^{d\times N}$, and the
outputs form the vector $\mathbf{y\in }$ $\mathbb{R}^{N}$. Furthermore, it
is assumed that%
\begin{equation}
y=f\left( \mathbf{q}\right) +\epsilon ,\quad \epsilon \sim \mathcal{N}%
(0,\sigma ^{2}),  \label{p1}
\end{equation}%
where $\epsilon $ is Gaussian noise with zero mean and variance $\sigma ^{2}$%
.

For low dimensionality $d$ of the inputs, linear regression cannot satisfy a
variety of dependencies encountered in practice. Therefore, the input is
mapped into a high-dimensional feature space $\mathbb{R}^{D},$ $D>d$, and
the linear regression model is applied in this space. To avoid any confusion
with the original feature space $\mathbb{R}^{d}$, this space will be
referred to as the ``extended feature space'' and its dimensionality can be
referred to as the ``size'' of the system. Denoting the mapping function $%
\mathbf{\phi }\left( \mathbf{q}\right) $, $\mathbf{\phi }:\mathbb{R}%
^{d}\rightarrow \mathbb{R}^{D}$ we have%
\begin{equation}
f\left( \mathbf{q}\right) =\mathbf{\phi }\left( \mathbf{q}\right) ^{T}%
\mathbf{w=}\sum_{j=1}^{D}w_{j}\phi _{j}\left( \mathbf{q}\right) \mathbf{%
,\quad w}=\left( w_{1},...,w_{D}\right) \in \mathbb{R}^{D}.  \label{p4}
\end{equation}%
The predictive mean and the covariance matrix can then be computed as%
\begin{equation}
\overline{\mathbf{w}}=\frac{1}{\sigma ^{2}}\Sigma \Phi \mathbf{y,\quad }%
\Sigma ^{-1}=\frac{1}{\sigma ^{2}}\Phi \Phi ^{T}+\Sigma _{p}^{-1}.
\label{p5}
\end{equation}%
The Bayesian formalism provides Gaussian distributions for the training
outputs and for the predictive mean $m_{\ast }$ and variance $v_{\ast }$.%
\begin{eqnarray}
\mathbf{y} &\sim &\mathcal{N}(0,\Sigma _{l}),\quad y_{\ast }\sim \mathcal{N}%
(m_{\ast },v_{\ast }),  \label{p6} \\
\Sigma _{l}^{-1} &=&\frac{1}{\sigma ^{2}}\left( I-\frac{1}{\sigma ^{2}}\Phi
^{T}\Sigma \Phi \right) ,  \nonumber \\
m_{\ast } &=&\mathbf{\phi }\left( \mathbf{q}_{\ast }\right) ^{T}\overline{%
\mathbf{w}},\quad v_{\ast }=\mathbf{\phi }\left( \mathbf{q}_{\ast }\right)
^{T}\Sigma \mathbf{\phi }\left( \mathbf{q}_{\ast }\right) +\sigma ^{2}, 
\label{p11}
\end{eqnarray}%
where $I$ is the identity matrix. These expressions allow for the
computation of the log-marginal likelihood (LML), 
\begin{equation}
\log p\left( \mathbf{y}|Q\right) =-\frac{1}{2}\mathbf{y}^{T}\Sigma _{l}^{-1}%
\mathbf{y}-\frac{1}{2}\log \left| \Sigma _{l}\right| -\frac{N}{2}\log \left(
2\pi \right) .  \label{p7}
\end{equation}%
Maximization of this function leads to optimal values of hyper-parameters
such as $\sigma ^{2}$ and other variables that define $\mathbf{\phi }\left( 
\mathbf{q}\right) $.

\section{Multiscale sparse GP regression}

As stated earlier, the goal of the proposed GP-based regression process is
to decrease the complexity of both training and testing and to make the
prediction more robust for datasets that have a highly irregular
distribution of features.

We consider Gaussian basis functions,%
\begin{equation}
\phi _{j}\left( \mathbf{q}\right) =g\left( \mathbf{q},\mathbf{q}_{j}^{\prime
},h_{j}\right) =\exp \left( -\frac{\left\| \mathbf{q-q}_{j}^{\prime
}\right\| ^{2}}{h_{j}^{2}}\right) ,\quad j=1,...,D,\quad \mathbf{\phi }%
=\left( \phi _{1},...,\phi _{D}\right) ^{T},  \label{m1}
\end{equation}%
where each function is characterized not only by its center $\mathbf{q}_{j}$%
, but now also by a scale $h_{j}$. The need for multiple scales may arise
from the underlying physics (e.g. particle density estimation) or from the
substantial non-uniformity of the input data distribution, which could, for
instance, demand smaller scales in denser regions. Note that the matrix $%
\Phi \Phi ^{T}$ in Eq. (\ref{p5}) is given by 
\begin{eqnarray}
\left( \Phi \Phi ^{T}\right) _{ij} &=&\sum_{n=1}^{N}\phi _{i}\left( \mathbf{q%
}_{n}\right) \phi _{j}\left( \mathbf{q}_{n}\right)   \label{m2} \\
&=&\sum_{n=1}^{N}\exp \left( -\frac{\left\| \mathbf{q}_{n}-\mathbf{q}%
_{i}^{\prime }\right\| ^{2}}{h_{i}^{2}}-\frac{\left\| \mathbf{q}_{n}-\mathbf{%
q}_{j}^{\prime }\right\| ^{2}}{h_{j}^{2}}\right) ,\quad i,j=1,...,D, 
\nonumber
\end{eqnarray}

\subsection{Sparse representations}

While $N$ training points may be available and maximum information can be
obtained when the size of the extended feature space $D=N$, we will search
for subsets of these points leading to a lower size $D<N$. The size of the
extended feature space is related to the accuracy of the low-rank
approximation of matrix $\Phi $ built on the entire training set (i.e. for
the case $D=N$).

Assume for a moment that we have a single scale, $h_{1}=...=h_{N}=h.$ If $h$
is chosen to be smaller than the minimum distance between the training
points, the matrix $\Phi $ will have a low condition number and a low
marginal likelihood. On the other end, if $h$ is substantially larger than
the maximum distance between the training points the problem will be
ill-posed and the marginal likelihood will be also low. The optima should
lie between these extremes. If the solution is not sensitive to the removal
of a few training points, a good low-rank approximation can be sought.

The fact that the $N\times N$ matrix $\Phi $ can be well-approximated with a
matrix of rank $r<N$ means that $N-r$ rows or columns of this matrix can be
expressed as a linear combination of the other rows or columns with
relatively small error $\epsilon ^{\prime }$. Assuming that $r$ locations
(or centers) of the radial basis functions (or Gaussians) are given and
denoting such centers by $\mathbf{q}_{j}^{\prime }$, $j=1,...,r$, we have 
\begin{equation}
\phi _{n}\left( \mathbf{q}\right) =g\left( \mathbf{q},\mathbf{q}%
_{n},h_{n}\right) =\sum_{j=1}^{r}\eta _{nj}\phi _{j}\left( \mathbf{q}\right)
+\epsilon _{n}^{\prime },\quad \mathbf{q}_{j}^{/}\in \mathbb{Q}^{\prime }%
\mathbb{\subset Q}\text{,\quad }n=1,...,N,  \label{m4}
\end{equation}%
where $\eta _{nj}$ are coefficients to be determined. Hence, output (Eqn. %
\ref{p1}), where $f$ is expanded as Eqn. \ref{p4} with $D=N$, can be written
as%
\begin{eqnarray}
y\left( \mathbf{q}\right) &=&\sum_{n=1}^{N}w_{n}\phi _{n}\left( \mathbf{q}%
\right) +\epsilon =\sum_{j=1}^{r}w_{j}^{\prime }\phi _{j}\left( \mathbf{q}%
\right) +\epsilon +\epsilon ^{\prime }\mathbf{,}  \label{m5} \\
w_{j}^{\prime } &=&\sum_{n=1}^{N}w_{n}\eta _{nj},\quad \epsilon ^{\prime
}=\sum_{n=1}^{N}w_{n}\epsilon _{n}^{\prime }.  \nonumber
\end{eqnarray}%
This shows that if $\epsilon ^{\prime }$ is lower or comparable with noise $%
\epsilon $, then a reduced basis can be used and one can simply set the size
of the extended feature space to be the rank of the low-rank approximation, $%
D=r$, and coefficients $w_{j}^{\prime }$ can be determined by solving a $%
D\times D$ system instead of an $N\times N$ system.

\subsection{Representative training subsets}

\label{3_2} Let us consider now the problem of determination of the centers
of the basis functions and scales $h_{j}$. 
If each data-point is assigned a different scale, then the optimization
problem will become unwieldy, as $D$ hyperparameters will have to be
determined. Some compromises are made, and we instead use a set of scales $%
h_{1},...,h_{S}$. In the limit of $S=1$, we have a single scale model, while
at $S=D$ we prescribe a scale to each training point. To reduce the number
of scales while providing a broad spectrum of scales, we propose the use of
hierarchical scales, e.g. $h_{s}=h_{1}\beta ^{s-1}$, $s=1,...,S$, where $%
h_{1}$ is of the order of the size of the computational domain and $\beta <1$
is some dimensionless parameter controlling the scale reduction.

While there exist several randomization-based strategies to obtain a
low-rank representation by choosing $D $ representative input points~\cite%
{quinonero2005}, we propose a more regular, structured approach. For a given
scale $h$, we can specify some distance $a=\gamma h$, where $\gamma <1$ is
chosen such that

\begin{itemize}
\item The distance from any input point to a basis point is less than $a$.
Such a construction provides an approximately uniform coverage of the entire
training set with $d$-dimensional balls of radius $a$.

\item The number of such balls is - in some sense - optimal. Note that
smaller $\gamma $ results in smaller errors $\epsilon _{n}^{\prime }$ and $%
\epsilon ^{\prime }$ and larger $r$ in Eqs. (\ref{m4}) and (\ref{m5}). If $%
\gamma <h/a_{\min }$, where $a_{\min }$ is the minimal distance between the
points in the training set, then $\epsilon^{\prime }=0$ and $r=D=N$, which
corresponds to a full rank representation.
\end{itemize}

The problem of constructing an optimal set as described above is
computationally NP-hard. However, approximate solutions can be obtained with
a modification of the well-known $k$-means algorithm~\cite{kmeans}. In
computational geometry problems, the $k$-means algorithm partitions a domain
in $d$-dimensional space into a prescribed number, $k$, of clusters. In the
present problem, the number of clusters is unknown, but the distance
parameter $a$ is prescribed. Thus, the number of centers depends on the
input point distribution and can be determined after the algorithm is
executed. Since only the cluster centers are required for the present
problem, the following algorithm is used: \vspace{0.7cm}

\begin{mdframed}
\center{\bf Algorithm \#1: Determination of cluster centers and k}
\vspace{0.3cm}

{\bf Input:} set of training points $\mathbb{Q=}\left\{ \mathbf{q}_{1},...,\mathbf{%
q}_{N}\right\} $, max cluster radius $a$.

\begin{enumerate}
\item Define $\mathbb{R}_{0}=\mathbb{Q}$ and set $k=0$;

\item Do steps 3 to 6 while $\mathbb{R}_{k}\neq \emptyset ;$

\item $k=k+1;$

\item Assign $\mathbf{q}_{k}^{\prime }=\mathbf{q}_{j}\in $ $\mathbb{R}_{k-1}$
(where $\mathbf{q}_{j}$ is a random point from set $\mathbb{R}_{k-1}$)

\item Find all points $\mathbf{q}_{ki}\in \mathbb{R}_{k-1}$, such that $%
\left| \mathbf{q}_{ki}-\mathbf{q}_{k}^{\prime }\right| \leqslant a$. 

\item Define $\mathbb{Q}_{k}=\left\{ \mathbf{q}_{ki}\right\} $ and $\mathbb{R%
}_{k}=\mathbb{R}_{k-1}\backslash \mathbb{Q}_{k}$.
\end{enumerate}

{\bf Output:} set of cluster centers $\mathbb{Q}^{\prime }\mathbb{=}\left\{ 
\mathbf{q}_{1}^{\prime },...,\mathbf{q}_{k}^{\prime }\right\} $, number of
clusters $k$.

\end{mdframed}
\vspace{0.3cm}

The construction of training subsets in the case of multiple scales requires
further modification of the algorithm. Starting with the coarsest scale $%
h_{1}$, $k_{1}$ centers can be determined using Algorithm 1 with distance
parameter $a_{1}=\gamma h_{1}$. In our approach, we select the bases such
that each input point can serve as a center of only one basis function;
therefore, the $k_{1}$ cluster centers of scale $h_{1}$ should be removed
from the initial training set to proceed further. Next, we partition the
remaining set using Algorithm 1 with distance parameter $a_{2}=\gamma h_{2}$
and determine $k_{2}$ cluster centers. After removal of these points we
repeat the process until scale $h_{S}$ is reached at which we determine $%
k_{S}$ cluster centers and stop the process. This algorithm is described
below.

\vspace{0.3cm} 
\begin{mdframed}

\center{\bf Algorithm \#2: Determination of cluster centers in multiple scales}
\vspace{0.3cm}

{\bf Input:} Set of training points $\mathbb{Q}=\left\{ \mathbf{q}_{1},...,\mathbf{%
q}_{N}\right\} $, set of scales $\mathbb{H=}\left\{ h_{1},...,h_{S}\right\} $, parameter $\gamma $.

\begin{enumerate}
\item Define $\mathbb{R=Q};$

\item Do steps 3 to 4 for $s=1,...,S;$

\item Execute Algorithm \#1 with input $\mathbb{R}$ and max cluster radius $%
a=\gamma h_{s}$ and get cluster centers $\mathbb{Q}_{s}^{\prime }$ and
number of clusters $k_{s};$

\item Set $\mathbb{R=R}\backslash \mathbb{Q}_{s}^{\prime }.$
\end{enumerate}

{\bf Output:} set of cluster centers $\mathbb{Q}^{\prime }=\mathbb{\cup Q}%
_{s}^{\prime }$, $\mathbb{Q}_{s}^{\prime }=\left\{ \mathbf{q}_{1}^{(s)\prime
},...,\mathbf{q}_{k_{s}}^{(s)\prime }\right\} $, number of clusters for each
scale $k_{s}$, $s=1,...,S$.

\end{mdframed}
\vspace{0.3cm}

\begin{figure}[tbp]
\centering \pic{0.4}{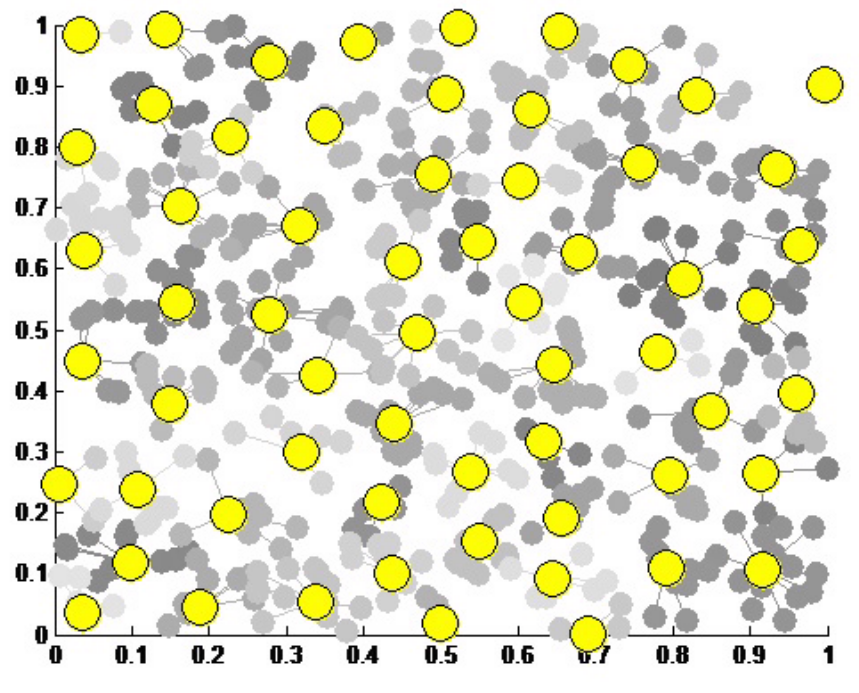}~~ \pic{0.4}{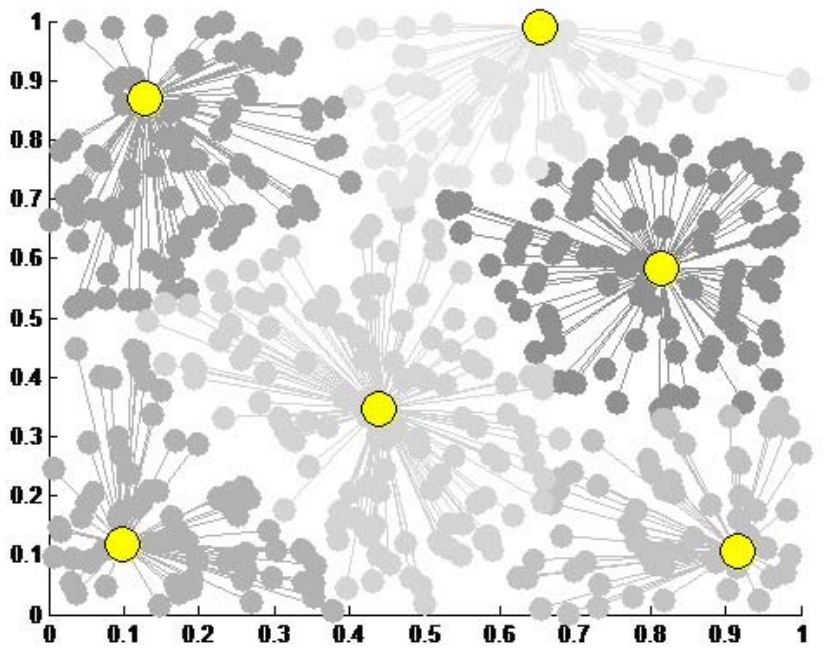}
\caption{The separation of 500 random points distributed inside a unit
square into clusters. The cluster centers are shown in yellow; points
belonging to different clusters are in different shades of gray. Left: small 
$h_n$. Right: large $h_n$.}
\label{mgp2d1}
\end{figure}

Figure \ref{mgp2d1} illustrates the clustering of a 2-dimensional dataset.
In principle, one can use the process even further until all the points in
the training set become the cluster centers at some scale (i.e. $D=N$).
However, as mentioned above, the reduction of the basis, $D<N$, is important
to reduce the computational complexity of the overall problem.

\section{Complexity}

The asymptotic complexity of algorithm \#1 is $O\left( Nk\right) $ and that
of algorithm \#2 is $O\left( ND\right) $. While fast neighbor search methods~%
\cite{cheng1984} can accelerate these algorithms, the overall computational
cost of the regression is still expected to be dominated by the training and
evaluation costs. It is thus instructive to discuss the complexities of
different algorithms available for training and testing, especially when $%
D\ll N$.

\subsection{Training}

The goal of the training stage is to determine quantities necessary for
computations in the testing stage. At this point, two types of computations
should be recognized: i) pre-computations, when the set of hyperparameters
is given, and ii) the optimization problem to determine the hyperparameters. 

To accomplish the first task, we use the following steps: 

\begin{itemize}
\item The initial step is to form the matrix $\Phi $, and then the $D\times D
$ matrix $\Sigma ^{-1}$ in Eq. (\ref{p5}). According to Eq. (\ref{m2}), the cost of this step is $O\left( ND^{2}\right) $. Note that this
cost is much larger than the $O\left( ND\right) $ required to determine the
basis function centers using the $k$-means type algorithm described in
Section \ref{3_2}.

\item The next step involves the determination of the weights $\overline{%
\mathbf{w}}$ from Eq. (\ref{p5}) This requires solving a $D\times D$. If
solutions are obtained through direct methods, the cost is $O\left(
D^{3}+ND\right) $.

\item Following this, the inverse $\Sigma $ should be determined\footnote{%
Alternatively, efficient decompositions enabling solutions with multiple
right hand sides can be used.}. The Cholesky decomposition is typically used
in this situation~\cite{rbfgp2}; 
\begin{equation}
\Sigma ^{-1}=L_{D}L_{D}^{T},  \label{m6}
\end{equation}%
where $L_{D}$ is the lower triangular matrices. The decompositions can be
performed with cost $O\left( D^{3}\right) $, respectively. Note that the
Cholesky decompositions can also be used to solve the systems for $\overline{%
\mathbf{w}}$, in which case the complexity of solution becomes $O\left(
D^{2}+ND\right) $.
\end{itemize}

The second task requires the computation of the log marginal likelihood. If
one uses Eq. (\ref{p7}) directly then the task becomes computationally very
expensive (complexity $O\left( N^{3}\right) $). The complexity can be
reduced with the aid of the following lemma.

\begin{lemma}
Determinants of matrices $\Sigma $ and $\Sigma _{l}$, defined by Eqs. (\ref%
{p5}) and (\ref{p6}), are related as%
\begin{equation}
\frac{1}{\sigma ^{2N}}\frac{\left| \Sigma \right| }{\left| \Sigma
_{p}\right| }=\frac{1}{\left| \Sigma _{l}\right| }.  \label{lm1}
\end{equation}

\begin{proof}
According to the definition of $\Sigma _{l}$ (\ref{p6}), we have in the
right hand side of Eq. (\ref{lm1}) 
\begin{equation}
\frac{1}{\left| \Sigma _{l}\right| }=\left| \Sigma _{l}^{-1}\right| =\frac{1%
}{\sigma ^{2N}}\left| I-\frac{1}{\sigma ^{2}}\Phi ^{T}\Sigma \Phi \right| .
\label{lm2}
\end{equation}%
Note now the Sylvester determinant theorem, which states that%
\begin{equation}
\left| I_{N}+AB\right| =\left| I_{D}+BA\right| ,  \label{lm3}
\end{equation}%
where $A$ is an $N\times D$ matrix and $B$ is $D\times N$, while $I_{N}$ and 
$I_{D}$ are the $N\times N$ and $D\times D$ identity matrices. Thus, we have
in our case%
\begin{equation}
\left| I-\frac{1}{\sigma ^{2}}\Phi ^{T}\Sigma \Phi \right| =\left| I_{N}-%
\frac{1}{\sigma ^{2}}\Phi ^{T}\left( \Sigma \Phi \right) \right| =\left|
I_{D}-\frac{1}{\sigma ^{2}}\left( \Sigma \Phi \right) \Phi ^{T}\right|
=\left| I-\frac{1}{\sigma ^{2}}\Sigma \Phi \Phi ^{T}\right| .  \label{lm4}
\end{equation}%
From Eq. (\ref{p5}) we have%
\begin{eqnarray}
I &=&\Sigma \Sigma ^{-1}=\Sigma \left( \frac{1}{\sigma ^{2}}\Phi \Phi
^{T}+\Sigma _{p}^{-1}\right) ,  \label{lm5} \\
\left| I-\frac{1}{\sigma ^{2}}\Sigma \Phi \Phi ^{T}\right| &=&\left| \Sigma
\Sigma _{p}^{-1}\right| =\frac{\left| \Sigma \right| }{\left| \Sigma
_{p}\right| }.  \nonumber
\end{eqnarray}%
Combining results (\ref{lm2}), (\ref{lm4}), and (\ref{lm5}), one can see
that Eq. (\ref{lm1}) holds.
\end{proof}
\end{lemma}

Now, replacing $\Sigma _{l}^{-1}\mathbf{y}$ in the first term in the
right-hand side of Eq. (\ref{p7}) with expressions for $\Sigma _{l}^{-1}$
and $\overline{\mathbf{w}}$ from Eqs. (\ref{p5}) and (\ref{p6}) and using
Eq. (\ref{lm1}) in the second term, we obtain 
\begin{equation}
\log p\left( \mathbf{y}|X\right) =-\frac{1}{2\sigma ^{2}}\left( \mathbf{y}%
-\Phi ^{T}\overline{\mathbf{w}}\right) ^{T}\mathbf{y}-\frac{1}{2}\log \left|
\Sigma ^{-1}\right| -\frac{1}{2}\log \left| \Sigma _{p}\right| -\frac{N}{2}%
\log \left( 2\pi \sigma ^{2}\right) .  \label{m8}
\end{equation}%
In the case of $\Sigma _{p}=\sigma _{p}^{2}I$ and using the Cholesky
decomposition of $\Sigma ^{-1}$, we obtain%
\begin{equation}
\log p\left( \mathbf{y}|X\right) =-\frac{1}{2\sigma ^{2}}\left( \mathbf{y}%
-\Phi ^{T}\overline{\mathbf{w}}\right) ^{T}\mathbf{y}-\sum_{j=1}^{D}\log
L_{D,jj}-\frac{N}{2}\log \left( 2\pi \sigma _{p}^{2}\sigma ^{2}\right) .
\label{m9}
\end{equation}%
Here, the cost of computing the first term on the right hand side is $%
O\left( ND\right) $, and the cost of the second term is $O\left( D\right) $.

Multi-dimensional optimization is usually an iterative process. Assuming
that the number of iterations is $N_{iter}$, we can write the costs of the
training steps marked by the respective superscripts as follows%
\begin{equation}
C_{train}=O\left( N_{iter}D\left( N+D^{2}\right) \right) \text{.}
\label{m10}
\end{equation}

\subsection{Testing}

The cost of testing can be estimated from Eq. (\ref{p6}) as the cost of
computing the predictive mean and variance. Taking into account the Cholesky
decomposition (\ref{m6}), these costs can be written as%
\begin{equation}
C_{mean}=O\left( MD\right) \text{,\quad }C_{var}=O\left( MD^{2}\right) \text{%
.}  \label{m12}
\end{equation}%
where $M$ is the number of test points. The cost for computing the variance
is much higher than the cost for computing the mean, because it is
determined by the cost of solving triangular systems of size $D\times D$ 
\emph{per} evaluation point.

\section{Numerical Results}
In this section, a set of simple analytical examples are first formulated to critically evaluate the accuracy and effectiveness of MGP and to
contrast its performance with conventional GP regression. Following this, data from a turbulent combustion simulation is used to 
assess the viability of the approach in scientific computing problems.
\subsection{Analytical examples}
\begin{figure}[ht]
	\centering
\includegraphics[width=0.95\textwidth]{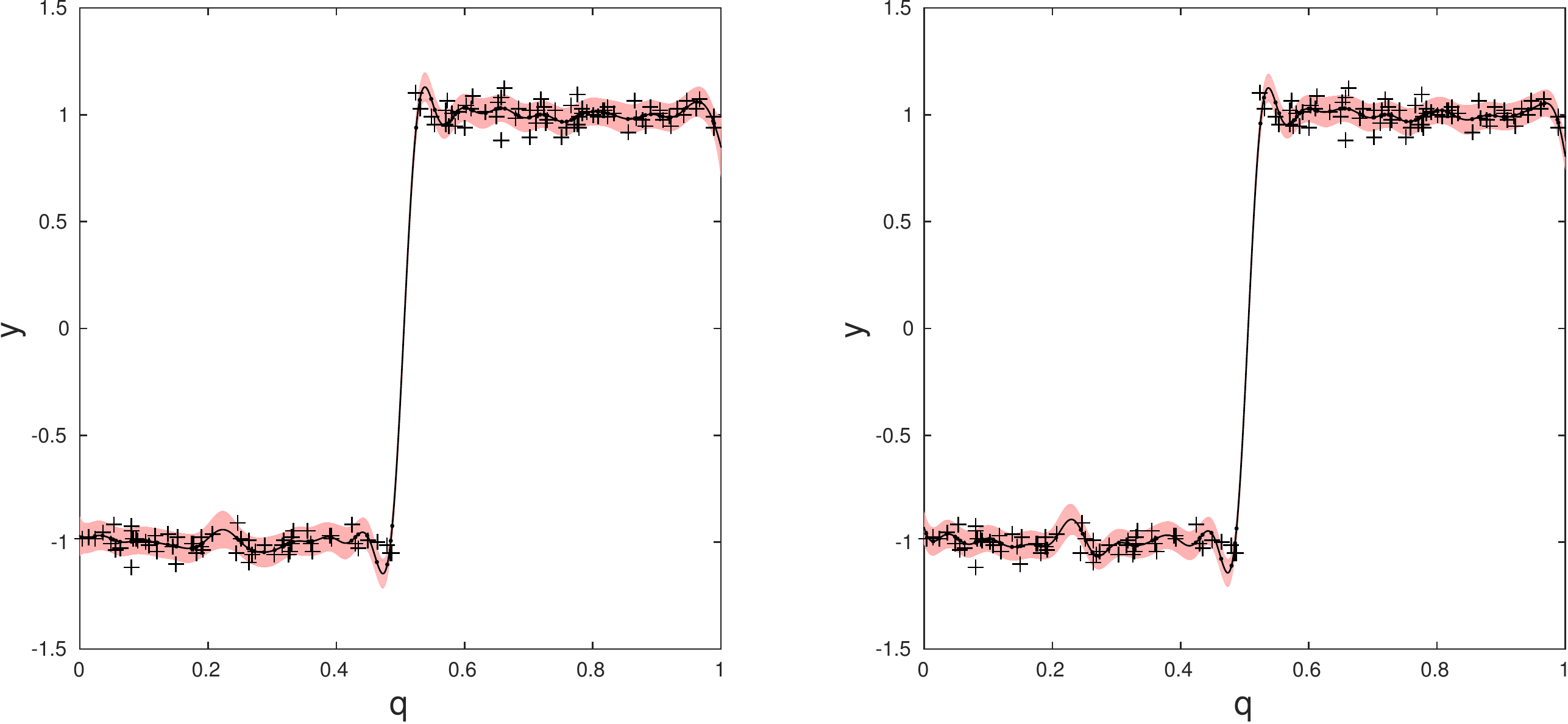}
\caption{Comparison of the output of the standard kernel GP using a Gaussian kernel (left)
and MGP with $S=1$ (right) for a step function. The crosses are training points ($N=128$), while the continuous lines and shaded
areas show the predictive mean and the variance.}  \label{stepfig1}
\end{figure}

\begin{figure}[ht]
	\centering
\includegraphics[width=0.95\textwidth]{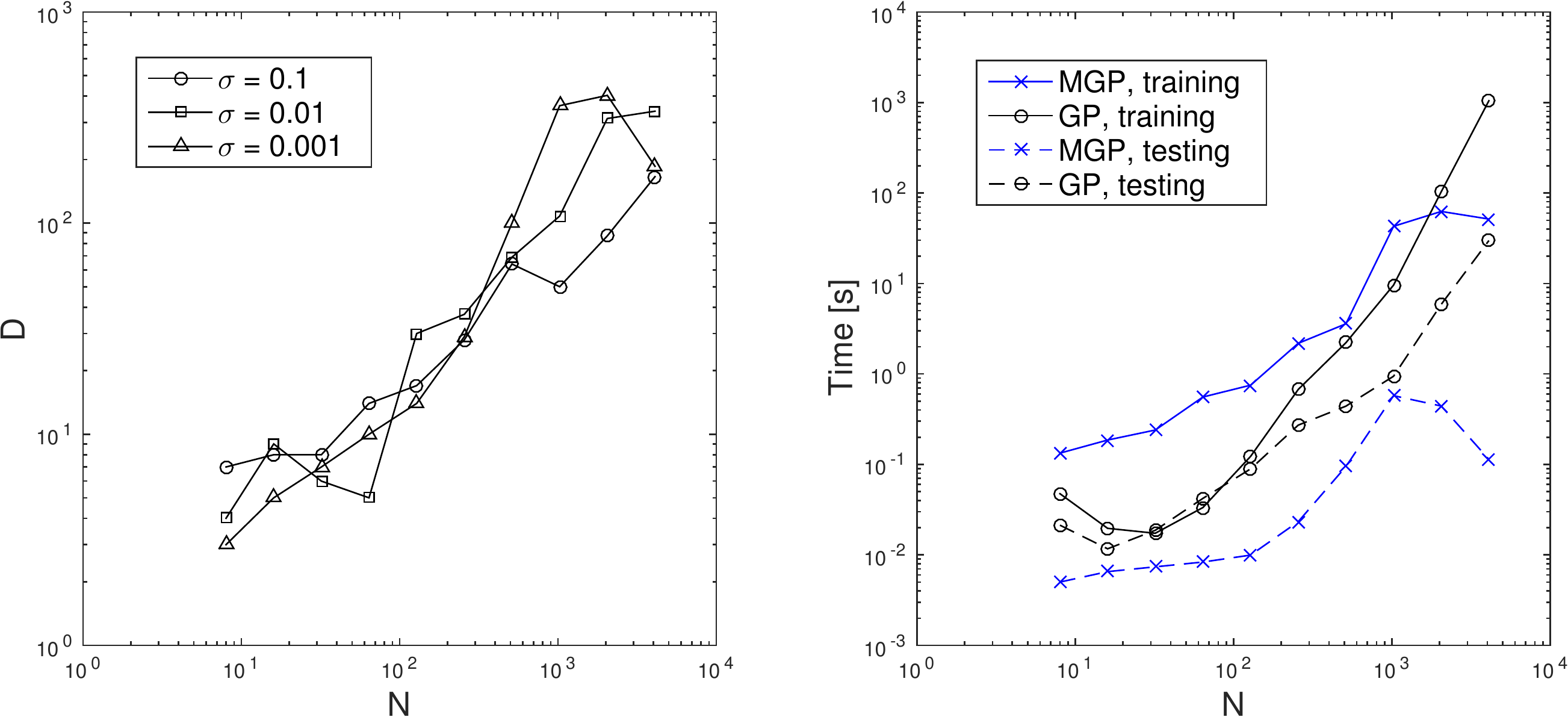}
\caption{Comparison of the performance of the standard GP and MGP for the step function in Figure \ref{stepfig1}. The plot on the left shows the size of the extended feature space for different input noise $\sigma$. The
wall clock time is obtained using MATLAB on a typical desktop PC.}  \label{stepfig1b}
\end{figure}

\begin{figure}[ht]
	\centering
\includegraphics[width=0.95\textwidth]{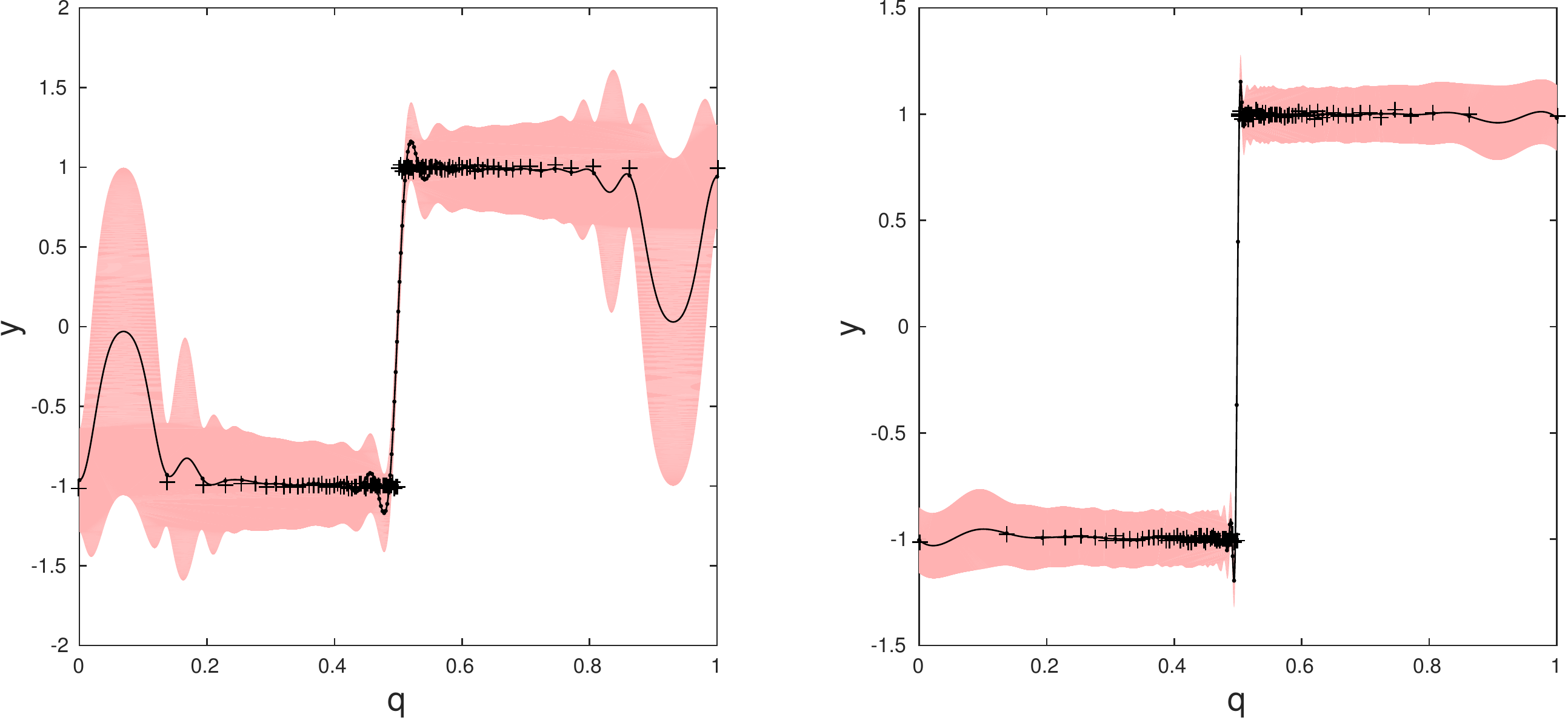}
\caption{Comparison of the standard GP (left) and MGP (right) for a step function. In contrast to Figure \ref{stepfig1}, the training point distribution is non-uniform with an exponential density increase near the jump. The number of training points $N=101$. Gaussian noise with $\sigma =0.03$
was added. The crosses
represent training points, while the continuous lines and shaded
areas show the predictive mean and the variance. The dots on the continuous
line have abscissas of the training points. }  \label{stepfig2}
\end{figure}

\begin{figure}[ht]
	\centering
\includegraphics[width=0.95\textwidth]{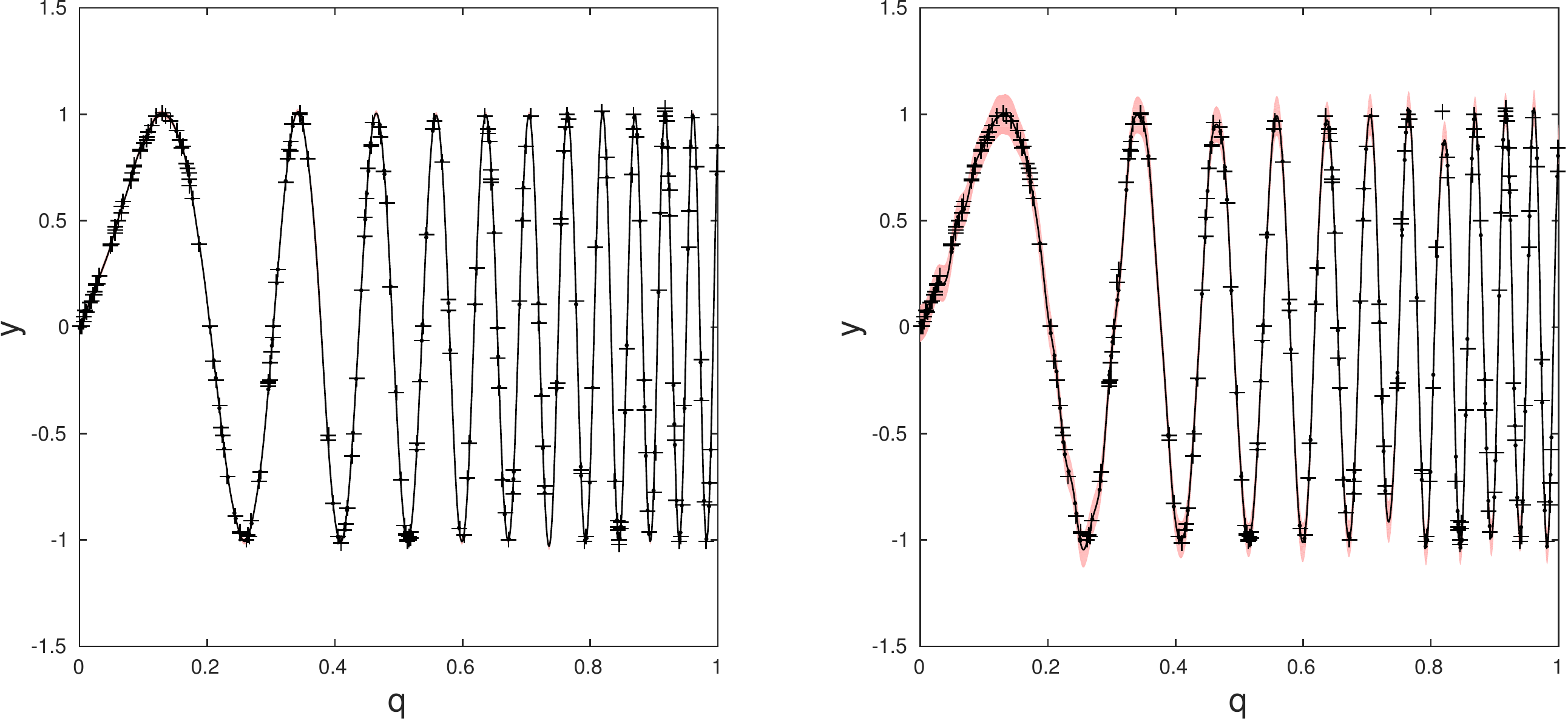}
\caption{Comparison of performance of the standard GP (left) and MGP with $S=1$ (right) for a sine function, $y = \sin(2\pi q(4q + 1)^{1.5})$ with Gaussian noise of standard deviation $\sigma = 0.01$. Crosses represent training points ($N=128$); continuous lines and shaded regions show the predictive mean and the variance.}  \label{sinefig1}
\end{figure}

\begin{figure}[ht]
	\centering
\includegraphics[width=0.95\textwidth]{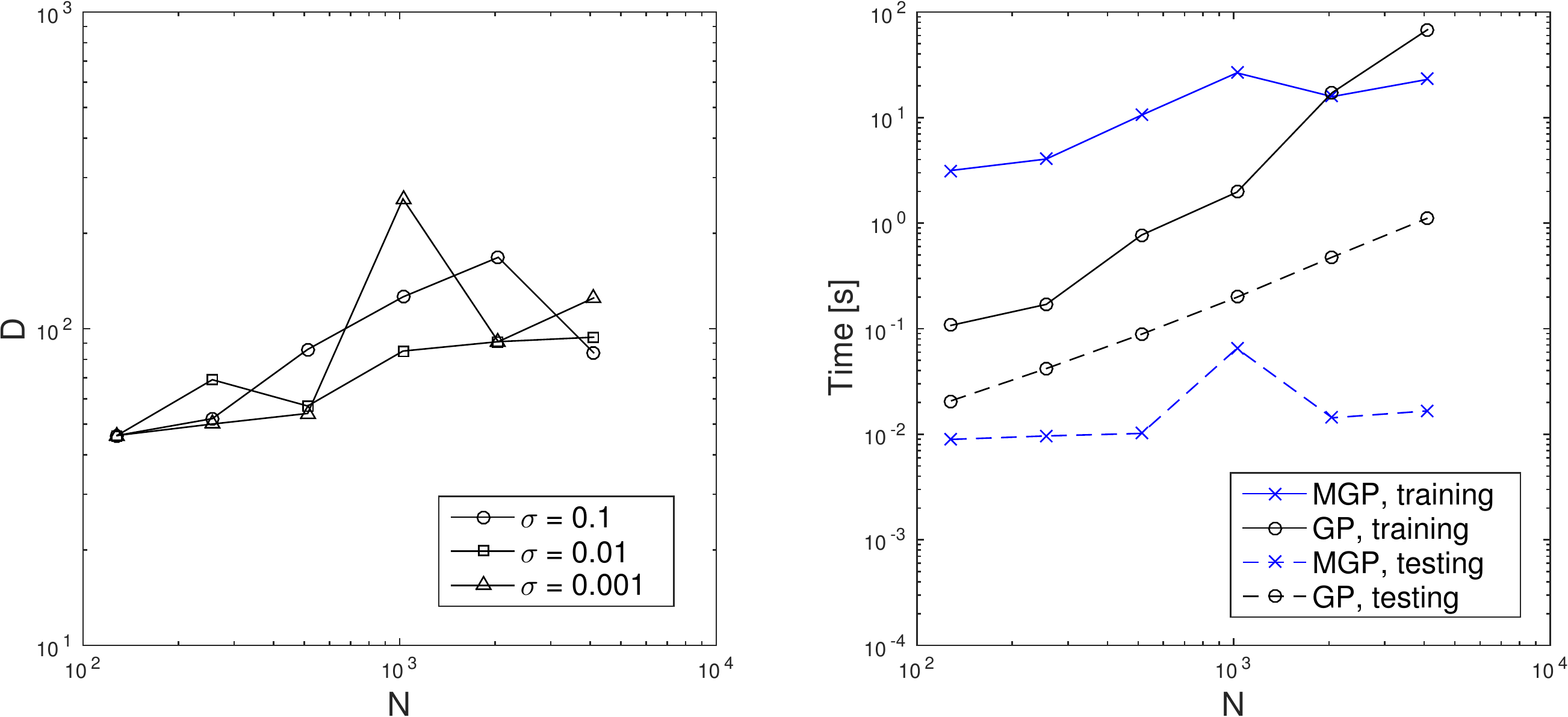}
\caption{Comparison of the performance of the standard GP and MGP for the sine function in Figure \ref{sinefig1}. The
wall-clock time is obtained using MATLAB on a typical desktop PC.}  \label{sinefig2}
\end{figure}

The first numerical example we present is a 1-D step function ($d=1$). We
compare the standard kernel GP with a Gaussian kernel and the present MGP
algorithm restricted to a single scale ($S=1$). The training set of size $N$ was randomly selected from 10000 points distributed
in a uniform grid. Points not used for training were used as test points. The initial data was
generated by adding normally distributed noise with standard deviation $%
\sigma$ to the step function. Optimal hyperparameters were found using the Lagarias algorithm for the Nelder-Mead method (as implemented by MATLAB's
\texttt{fminsearch} function) \cite{lagarias}. For the standard GP, $\sigma$
and $h$ were optimized, while for the MGP algorithm, the optimal $\gamma 
$ was determined in addition to $\sigma$ and $h$. In both cases, the optimal $\sigma$ was close  to the
actual $\sigma $ used for data generation. The ratio between the optimal $h$ for the kernel and the finite-dimensional
(``weight-space'') approaches was found to be approximately 
$1.4$, which is consistent with the difference of $\sqrt{2}$ predicted by
the theory (the distinction between the weight space and function space approaches is provided in Ref.~\cite{rbfgp2}). 

The plots in Figure \ref{stepfig1} show that there is no substantial
difference in the mean and the variance computed using both methods, while
the sparse algorithm required only $D=38$ functions compared to the $N=128$
required for the standard method. Figure~\ref{stepfig1b} shows the relationship between the extended feature space (D) with respect to the size of the training set (N).
 Ideally, this should be a linear dependence, since the step function is scale-independent. Due to random noise and the possibility that the optimization may
converge to local minima of the objective function, however, the relationship is not exactly linear. 
Since $D$ is nevertheless several times smaller than $N$, the wall-clock time for the present
algorithm is shorter than that for the standard GP in both testing and training, per optimizer iteration.
The total training time for the present algorithm can be larger than that of the 
standard algorithm due to the overhead associated with the larger number of hyperparameters and the resultant increase in the number of optimization iterations required. 
However, we observed this only for relatively small values of $N$. For larger $N$, such as $N=4096$, the present
algorithm was approximately 5, 10, and 20 times faster than the standard
algorithm for $\sigma =0.1$, $0.01$, and $0.001$, respectively.

Figure \ref{stepfig2} illustrates a case where the training points are distributed non-uniformly. Such situations
frequently appear in practical problems, where  regions of high functional gradients are sampled with higher density to provide a good representation of the function. For example, adaptive meshes to capture phenomena such as shockwaves and boundary layers in fluid flow fall in this category. 
In the case illustrated here, the training points were
distributed with exponential density near the jump ($q=0.5\pm h_{t}\ln z$, $%
\exp \left( -0.5/h_{t}\right) \leqslant z\leqslant 1$, where $z$ are
distributed uniformly at the nodes of a regular grid. $h_{t}=0.1$ was used). For MGP, the number of scales was $S=6$
and the other hyperparameters were optimized using the same routine as before. Due to multiple extrema in the objective function, it is rather difficult to optimize the number of \ scales, $S$. In practice, one should start from several initial guesses or fix some parameters such as $S$. We used several values of $%
S$ and observed almost no differences for $5\leq S \leq 10$,
while results for $S=1$ and $S=2$ were substantially different from the cases where $S>2$.

It is seen that the MGP provides a much better fit of the step
function in this case than the standard method. This is achieved due to its
broad spectrum of scales. In the present example, we obtained the following
optimal parameters for scales distributed as a geometric progression, $%
h_{s}=h_{1}\beta ^{s-1}$: $h_{\max }=h_{1}=0.1233$, $h_{\min }=h_{6}=0.0032$%
, and $\beta =0.4806$. Other optimal parameters were $\gamma =0.258$ and $\sigma
=0.127$. For the standard GP, the optimal scale was $h=0.0333$. Figure \ref{stepfig2}
shows that with only a single intermediate scale, it is impossible to
approximate the function between training points with a large spacing, whereas MGP provides a much better approximation.
Moreover, since $h_{\min }<h$, we also have a better approximation of the jump, i.e. of
small-scale behavior. This is clearly visible in the figure; the jump for
the standard GP is stretched over about 10 intervals between sampling
points, while the jump for MGP only extends over 3 intervals.
Note that in the present example, we obtained $D=N=101$, so the
wall-clock time for testing is not faster than the standard GP. However, this case
illustrates that multiple scales can provide good results for substantially
non-uniform distributions where one scale description is not sufficient. 

As final analytical example, we explore a sine wave with varying frequency, depicted in Figure \ref{sinefig1}. As before, $N$ training points are randomly chosen from a uniform distribution, and $M=10000-N$ test points are used. For MGP, $S=1$ was used. Compared to the previous examples, this is an intermediate case in terms of scale dependence. One noteworthy result from this dataset is that $D$ is almost constant with respect to $N$, as seen in Figure \ref{sinefig2}. This shows that when the function is relatively smooth, the optimization process is not limited to producing a linear relationship between $D$ and $N$. Another result is that the output variance of the multiscale method is visibly higher than that of the standard method. For the previous cases, the variances have been either been close to equal, or the standard method would produce the higher variance. This could be due to the fact that $h$ and $D$ are inherently related for the current method, whereas $h$ is unrestricted for the standard GP. Since $D<N$, the multiscale $h$ for $S=1$ is typically greater than the standard method's $h$. According to Eq. (\ref{p11}), this would result in greater variance. 

\subsection{Data from Turbulent Combustion}
Combustion in the presence of turbulent flow involves an enormous disparity in time and length scales, meaning that direct numerical simulations (DNS) are not possible
in practical problems. Large eddy simulations (LES), in which the effect of scales smaller than the mesh resolution ({\em i.e.} subgrid scales) 
are modeled, is often a pragmatic choice. A key difficulty in LES of combustion is the modeling of the subgrid scale fluxes 
in the species transport equations~\cite{kant1,kant2}. These terms
arise as a result of the low-pass filtering - represented by the operator $(\bar{\cdot})$ - of the governing equations, and are of the form 
\begin{equation}
f_k = \overline{\rho u_k C}-\frac{\overline{\rho u_k}\overline{\rho C}}{\bar{\rho}}
\end{equation}
where $\rho, u,C$ represent density, velocity and species mass fraction, respectively. 
Subgrid-scale closures based on concepts from non-reacting flows, such as the equation below,\footnote{$C_s, Sc$ are typically constants, and $S_{ij} = \frac{1}{2}\left[ 
\frac{\partial u_i}{\partial x_j} + \frac{\partial u_j}{\partial x_i} \right]$ is the strain-rate tensor. The superscript $(\tilde{.})$ denotes Favre-filtering
and is defined by $\tilde{q}=\frac{\overline{\rho q}}{\bar{q}}$. $\Delta$ is the filter size.}
\begin{equation}
f_{k} = -\frac{\overline{\rho} C_s^2 \Delta^2 \sqrt{2 \tilde{S}_{ij} \tilde{S}_{ij}}}{Sc}\frac{\partial \tilde{c}}{\partial x_k}
\end{equation} 
are found to be inadequate for turbulent combustion. Modeling of the scalar fluxes thus continues to be an active area of research, and many analytical models are being evaluated by the community. Reference ~\cite{chakraborthy} provides a concise summary of such developments in the area of premixed turbulent flames.
\begin{figure}[ht]
	\centering
	\pic{0.8}{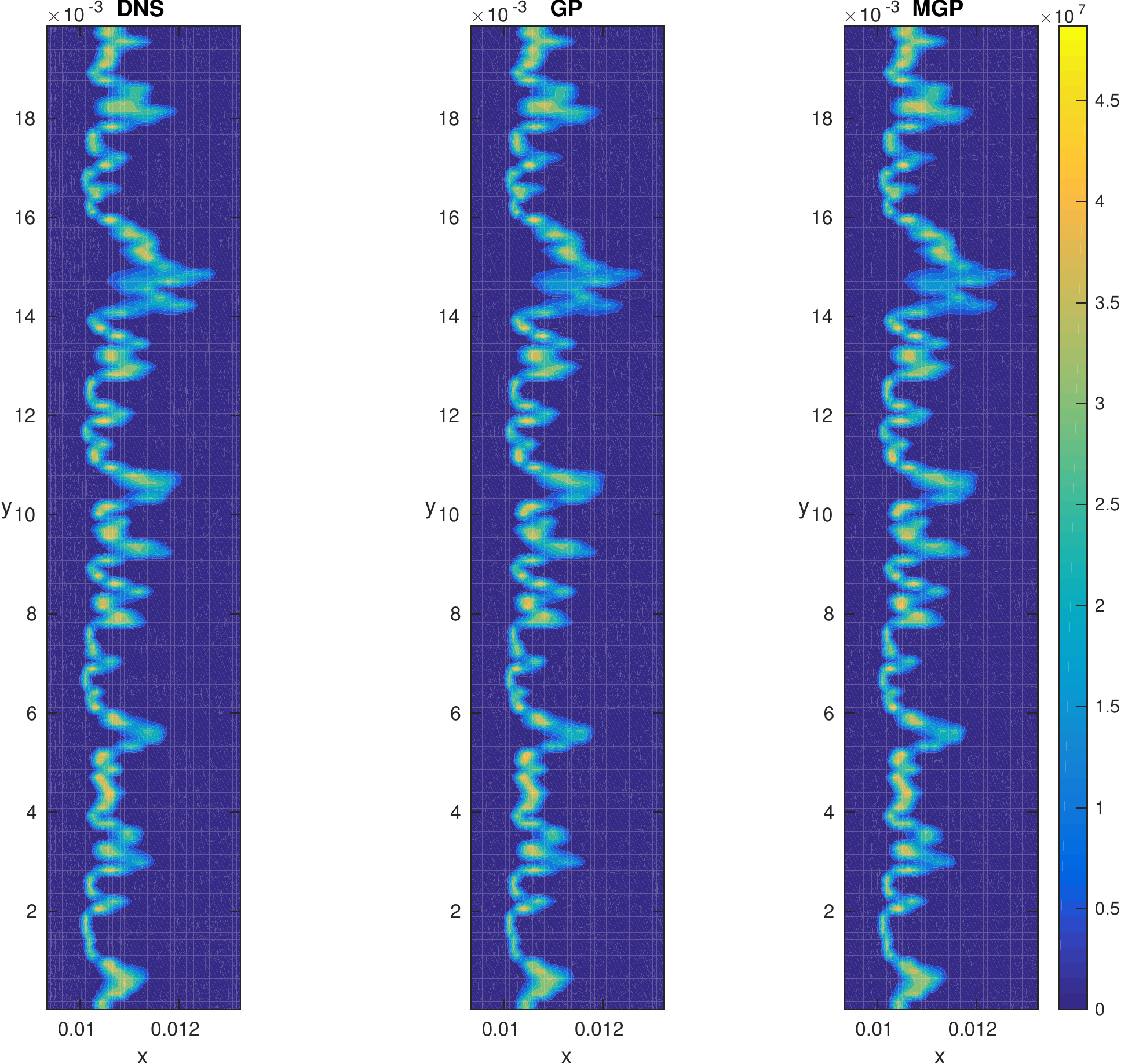}
		\caption{Contours of $F_1$ on the test $x-y$ plane. Left: DNS values (exact). Center: GP output. Right: MGP output.}  \label{combcontour}
\end{figure}

\begin{figure}[ht]
	\centering
	\pic{0.95}{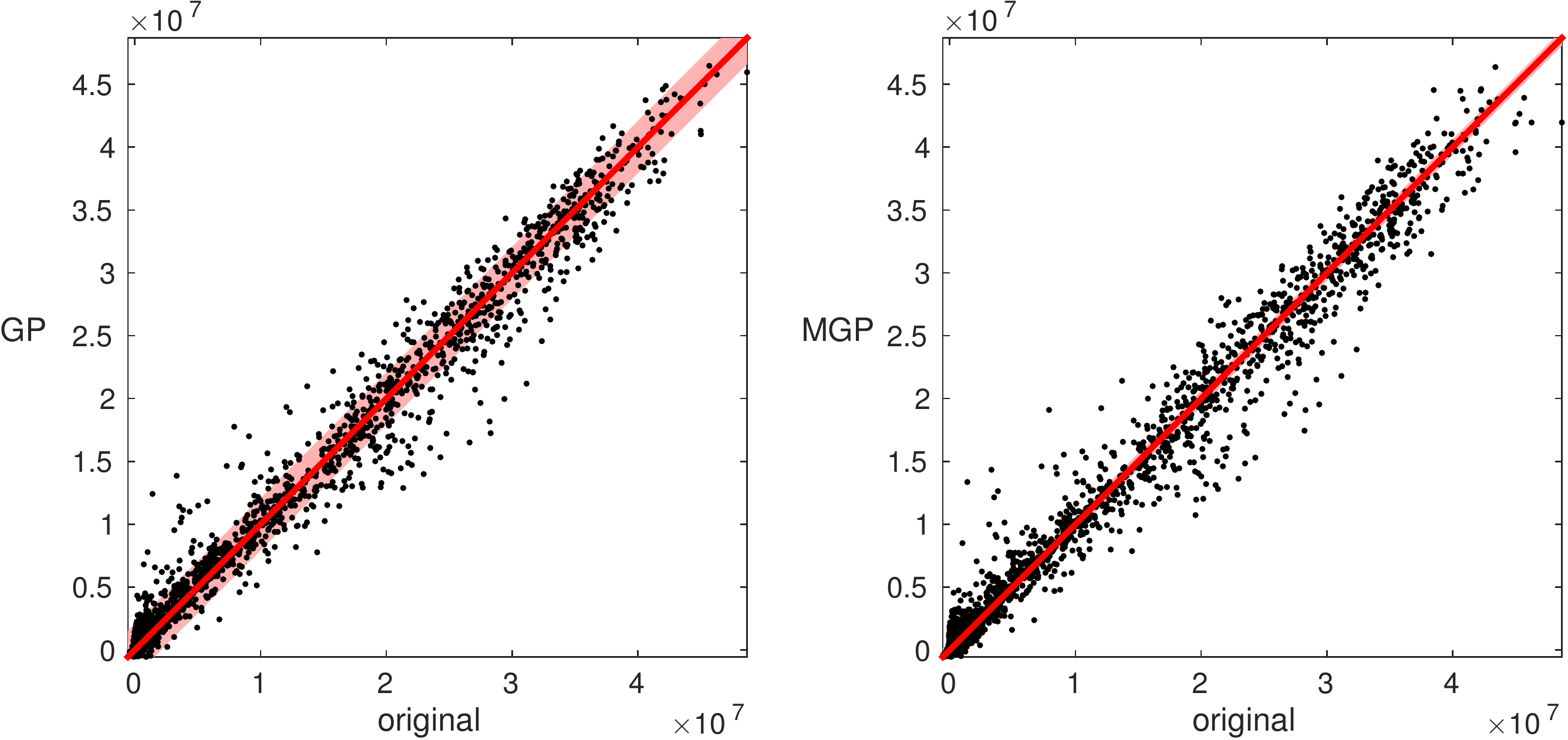}
		\caption{Original test output versus learned result for $F_1$. Red regions are $\pm\sigma$ from the diagonal, i.e. the optimized estimate of the input noise.}  \label{combscatter}
\end{figure}


In this work, we intend to apply GP and MGP to identify the model relationships from data. The simulation data is obtained from Ref.~\cite{raman}, in which
DNS of a propagating turbulent flame is performed. The configuration involves a premixed hydrogen-air flame propagating in the 
x-direction. The flame forms a thin front between the burnt and unburnt gases.
Specifically, we attempt to learn the normalized flux in the x-direction, $F_1$, as a function of seven variables:
\begin{equation}
F_1 = \frac{f_1}{\overline{\rho} \Delta^2} = f(\nabla \tilde{u},\nabla \tilde{c},\tilde{S})
\end{equation}

A total of 3 million training points were generated from the dataset by performing low-pass filtering.  Some care was taken in choosing training points. Since the flame is thin relative to the size of the domain, the majority of the data points were found to lie outside the region where $F_1$ is nonzero. To mitigate this disadvantageous distribution, 80 percent of the training points were randomly chosen from the data with $f_1 > 0.05$, and 20 percent were chosen from data with $f_1 \leq 0.05$, i.e. outside the flame. 45000 training points were used in total. For testing, a single $x-y$ plane of around 6500 points was set aside. 

The predictive results on this plane are shown in Table \ref{comberr}, and Figure \ref{combscatter} shows the ML output versus the true DNS values. From these plots, it is seen that MGP has achieved a fifty-fold increase in evaluation speed for a corresponding two percent increase in error. In the scatter plot, MGP appears to perform better than GP for low values of $F_1$ and worse for high values, but the overall difference is small. Figure \ref{combcontour} is a side-by-side comparison of contours of $F_1$ from DNS and from GP and MGP. It is especially evident in the contour plot that GP and MGP are both able to capture features of the flame, whereas analytical models described in Ref.~\cite{chakraborthy} were not as accurate\footnote{These results are not shown}.
 Figure \ref{combslice} plots $F_1$ along several locations perpendicular ($x$) and parallel ($y$) to the flame in the test plane. 

\begin{table}
\caption{Numerical results for learning $F_1$ in the flame. Error is defined as $||\mb{y} - f(\mb{Q}_*)||/||\mb{y}||$, where $\mb{y}$ is a vector of the exact outputs $F_1$, and $f(\mb{Q}_*)$ is a vector of the GP or MGP outputs at the test points $\mb{Q}_*$. Time is test time in seconds, obtained in MATLAB. MGP utilized 3 scales.}
\label{comberr}
\centering
  \begin{tabular}{| l | c | c | c | c |}
    \hline
    & \multicolumn{2}{c|}{GP} & \multicolumn{2}{c|}{MGP} \\ \hline
     & Error & Time & Error & Time \\ \hline
    $F_1$ & 0.1087 & $4.2\times10^1$ & 0.1105 & $8.0\times10^{-1}$ \\ 
    \hline
  \end{tabular}
\end{table}

\begin{figure}[ht]
\centering
	    \subfigure[$y=2\times10^{-5}$]{\includegraphics[width=80mm]{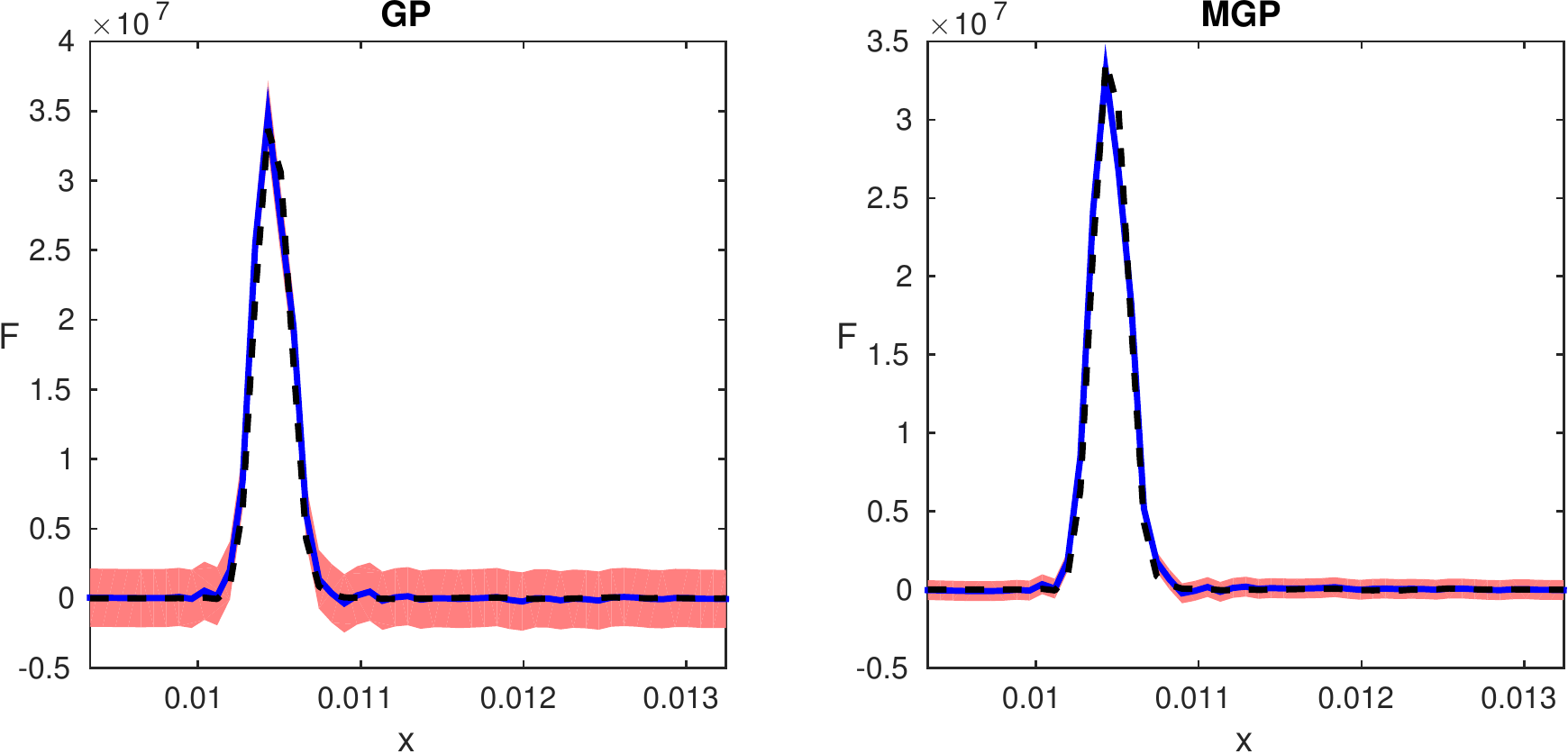}}
	    \subfigure[$y=8\times10^{-4}$]{\includegraphics[width=80mm]{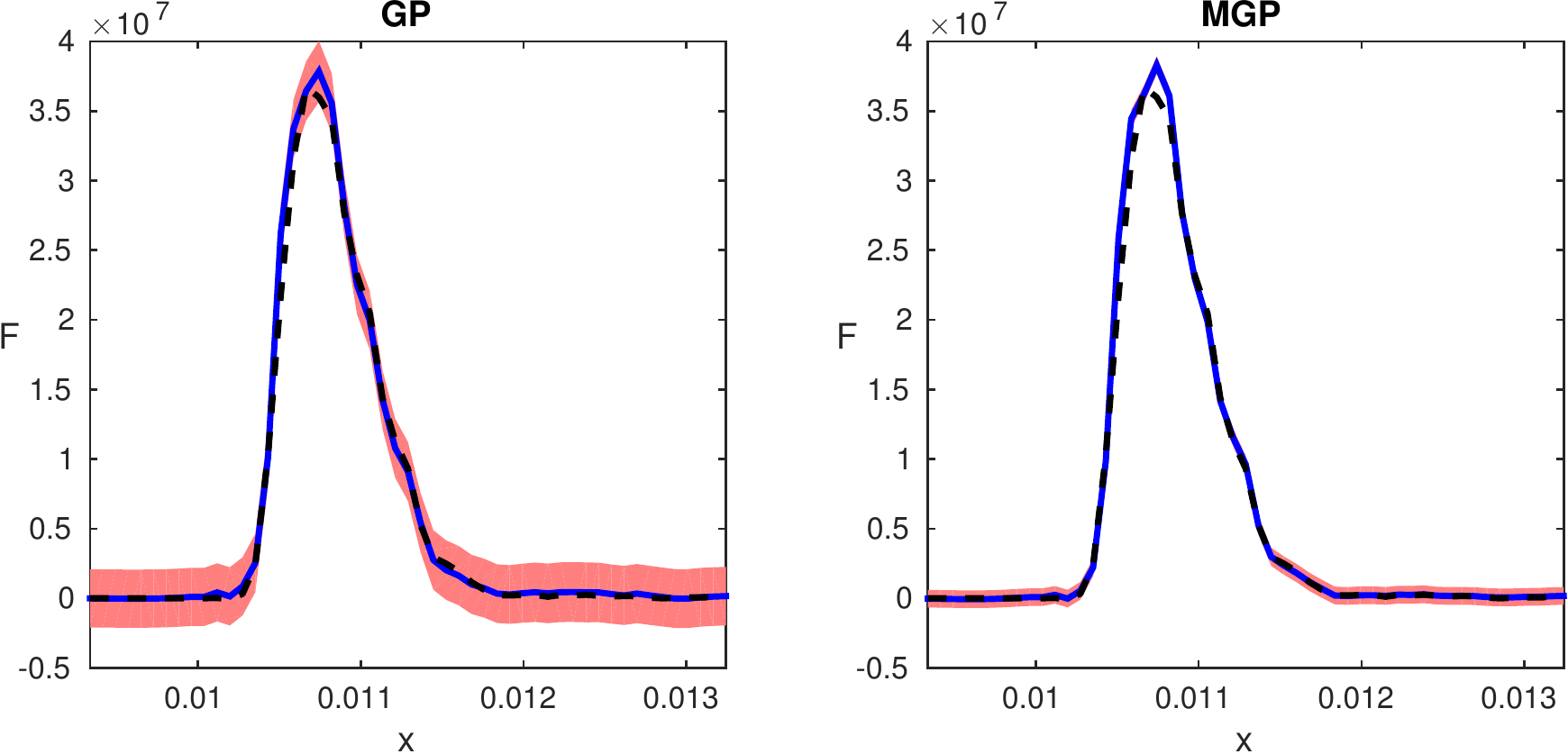}}
	    \subfigure[$y=5\times10^{-3}$]{\includegraphics[width=80mm]{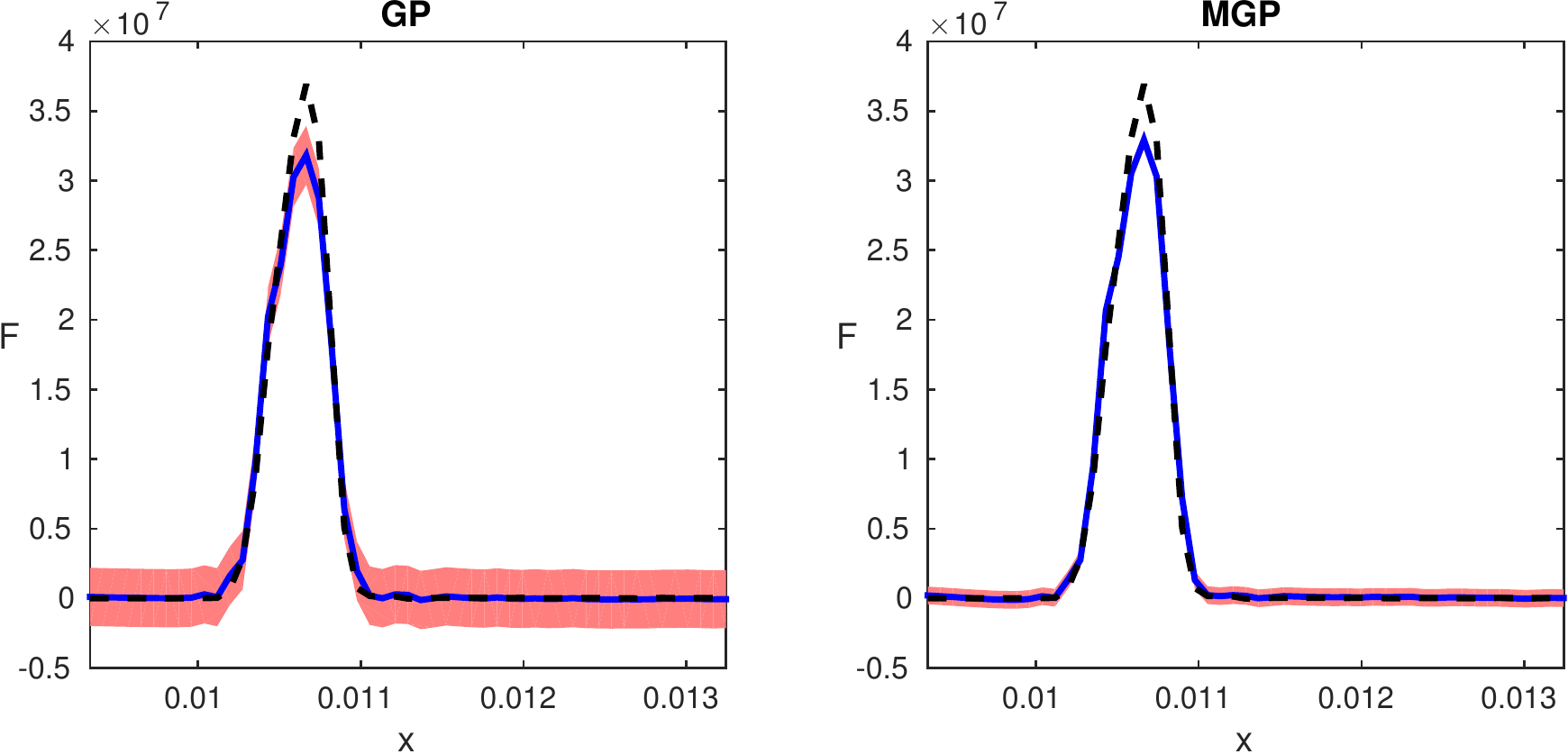}}
	    \subfigure[$y=1\times10^{-2}$]{\includegraphics[width=80mm]{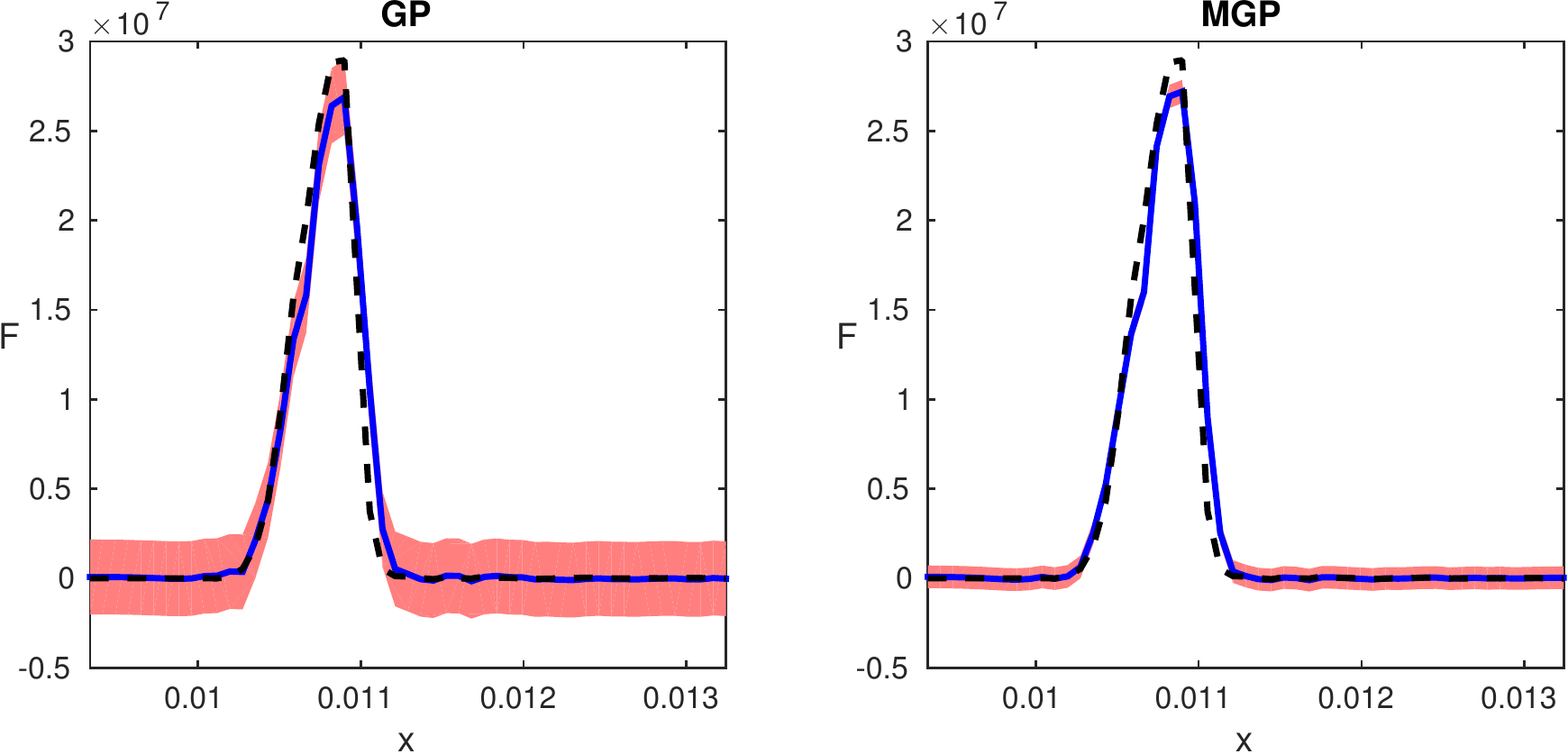}}
		\caption{$F_1$ as a function of $x$ for four different constant values of $y$ on the test $x-y$ plane. Dashed black lines are from DNS. GP results on the left, MGP results on the right. Red regions are one standard deviation for the output.}  \label{combslice}
\end{figure}

\section{Conclusion}
In this paper, a new Gaussian process (GP) regression technique was presented. The method, referred to as MGP, introduces multiple scales among the Gaussian basis functions and employs hierarchical clustering to select centers for these sparse basis functions. These modifications reduce the computational complexity of GP regression and also achieve better accuracy than standard GP regression when training points are non-uniformly distributed in the $d$-dimensional feature space. We illustrated these improvements through analytical examples and in a turbulent combustion datasets. 

In analytical examples with smooth functions, the MGP was shown to be at least an order of magnitude faster than the standard GP for a similar level of accuracy. In problems with discontinuities, the MGP is shown to provide a much better fit. 
In the turbulent combustion example in 7 dimensions, the MGP was shown to achieve a fifty-fold increase in evaluation speed for a corresponding two percent increase in error over the GP method. Overall, MGP is well-suited for regression problems where the inputs are unevenly distributed or where training and testing speeds are critical. 

Based on the results presented in this work, MGP offers promise as a potentially attractive method for use in many scientific computing applications in which datasets may be large, and sparsely distributed in a high-dimensional feature space. The MGP can be especially useful when predictive evaluations are performed frequently. However, further developments and more detailed application studies are warranted. 
It was observed that the optimization process is more likely to terminate at local minima compared to conventional GPs. 
An immediate area of investigation could 
explore more efficient and robust techniques for optimization of the hyperparameters. 
Since an appealing feature of MGP is the reduced complexity when working with large datasets, efficient parallelization strategies should be explored. 

The software and all the examples in this paper are openly available at \url{http://umich.edu/~caslab/#software}.


\section*{Acknowledgments}
This work was performed under the NASA LEARN grant titled ``A Framework for Turbulence Modeling Using Big Data.''

\bibliography{refs}
\bibliographystyle{aiaa}

\section{Numerical Results}
In this section, a set of simple analytical examples are first formulated to critically evaluate the accuracy and effectiveness of MGP and to
contrast its performance with conventional GP regression. Following this, data from a turbulent combustion simulation is used to 
assess the viability of the approach in scientific computing problems.
\subsection{Analytical examples}
\begin{figure}[ht]
	\centering
\includegraphics[width=0.95\textwidth]{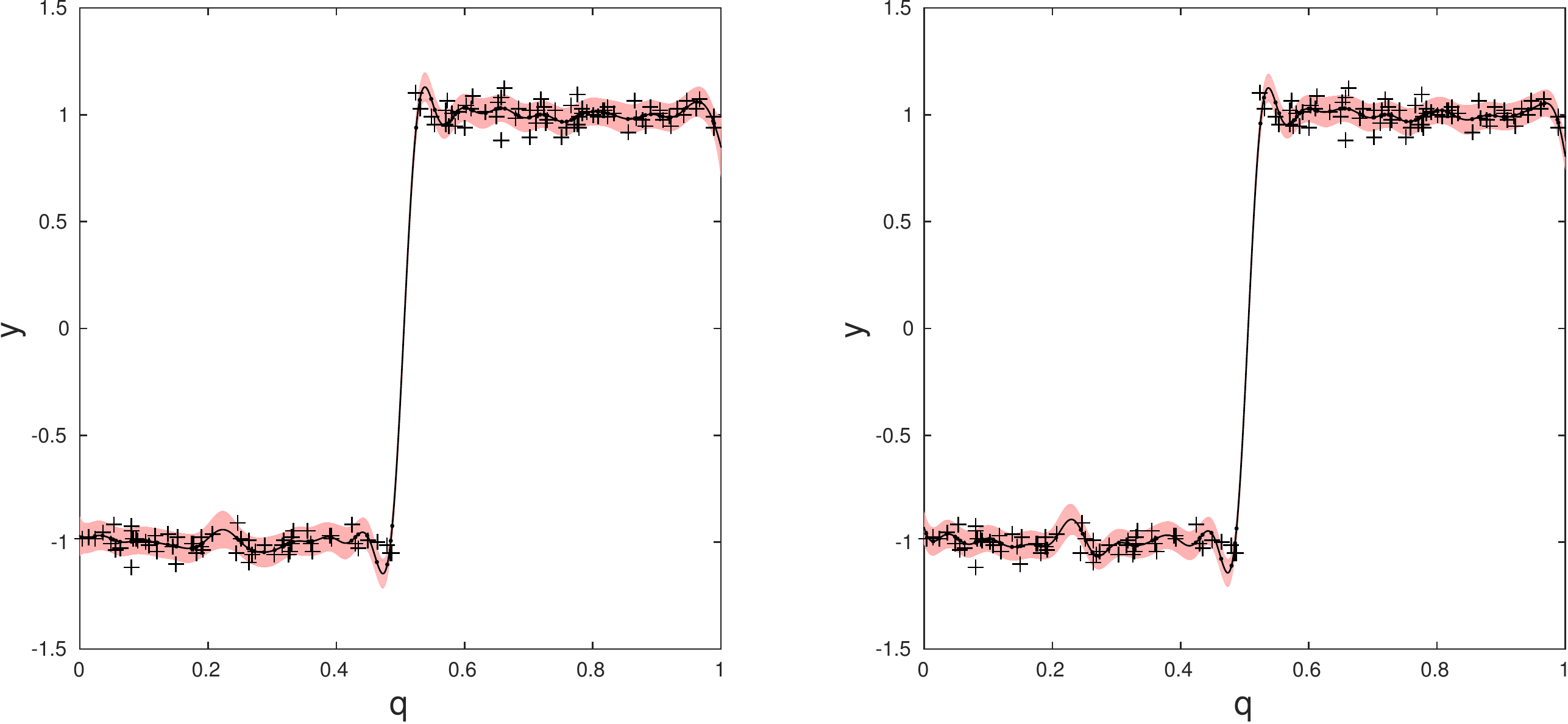}
\caption{Comparison of the output of the standard kernel GP using a Gaussian kernel (left)
and MGP with $S=1$ (right) for a step function. The crosses are training points ($N=128$), while the continuous lines and shaded
areas show the predictive mean and the variance.}  \label{stepfig1}
\end{figure}

\begin{figure}[ht]
	\centering
\includegraphics[width=0.95\textwidth]{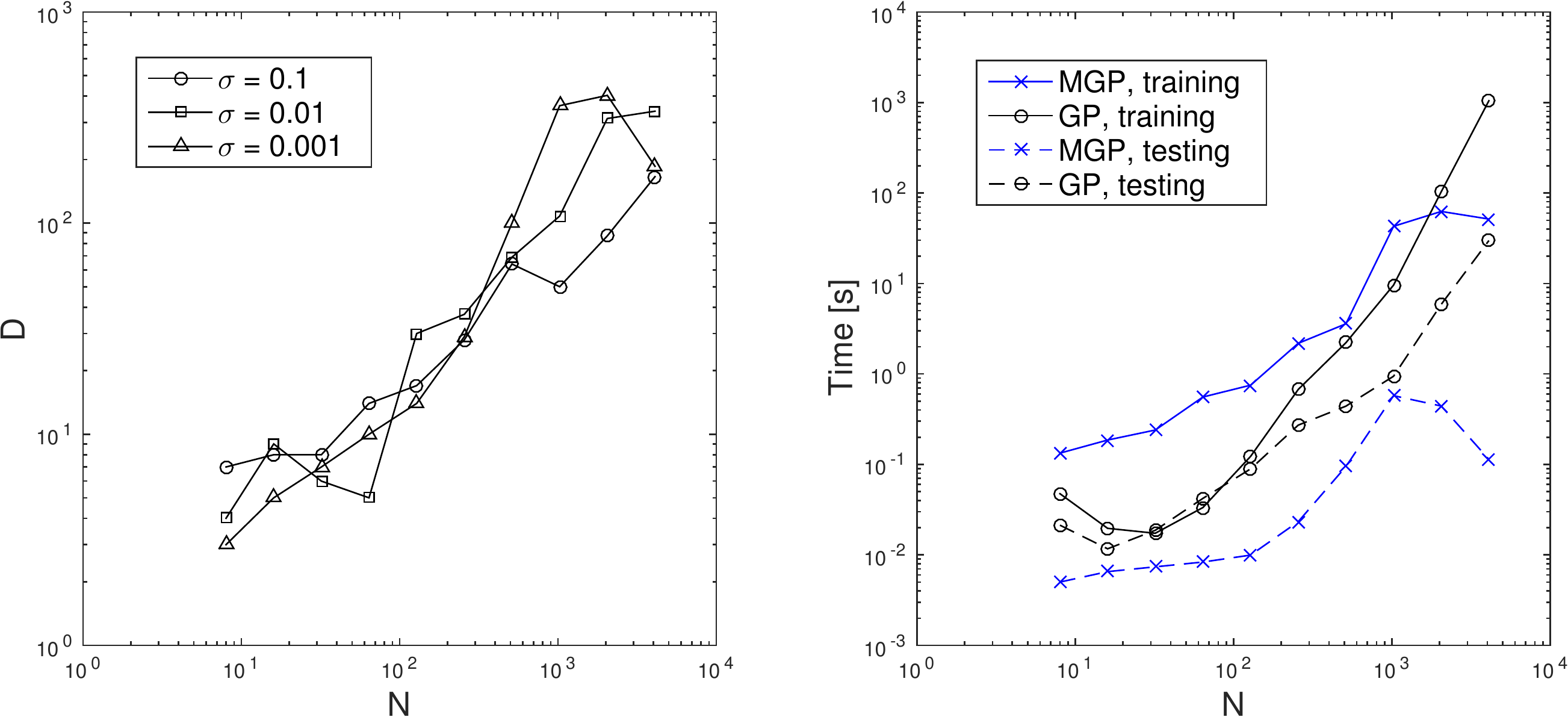}
\caption{Comparison of the performance of the standard GP and MGP for the step function in Figure \ref{stepfig1}. The plot on the left shows the size of the extended feature space for different input noise $\sigma$. The
wall clock time is obtained using MATLAB on a typical desktop PC.}  \label{stepfig1b}
\end{figure}

\begin{figure}[ht]
	\centering
\includegraphics[width=0.95\textwidth]{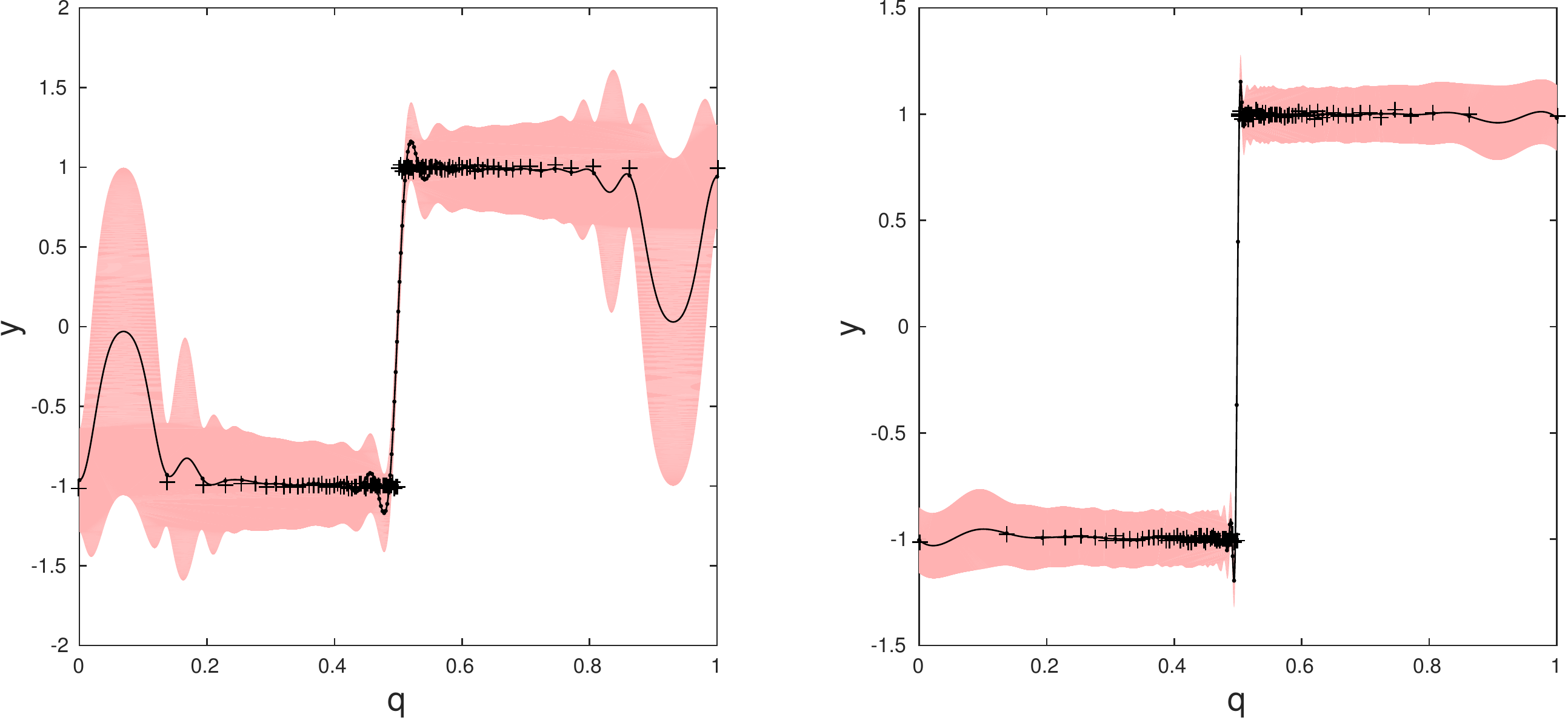}
\caption{Comparison of the standard GP (left) and MGP (right) for a step function. In contrast to Figure \ref{stepfig1}, the training point distribution is non-uniform with an exponential density increase near the jump. The number of training points $N=101$. Gaussian noise with $\sigma =0.03$
was added. The crosses
represent training points, while the continuous lines and shaded
areas show the predictive mean and the variance. The dots on the continuous
line have abscissas of the training points. }  \label{stepfig2}
\end{figure}

\begin{figure}[ht]
	\centering
\includegraphics[width=0.95\textwidth]{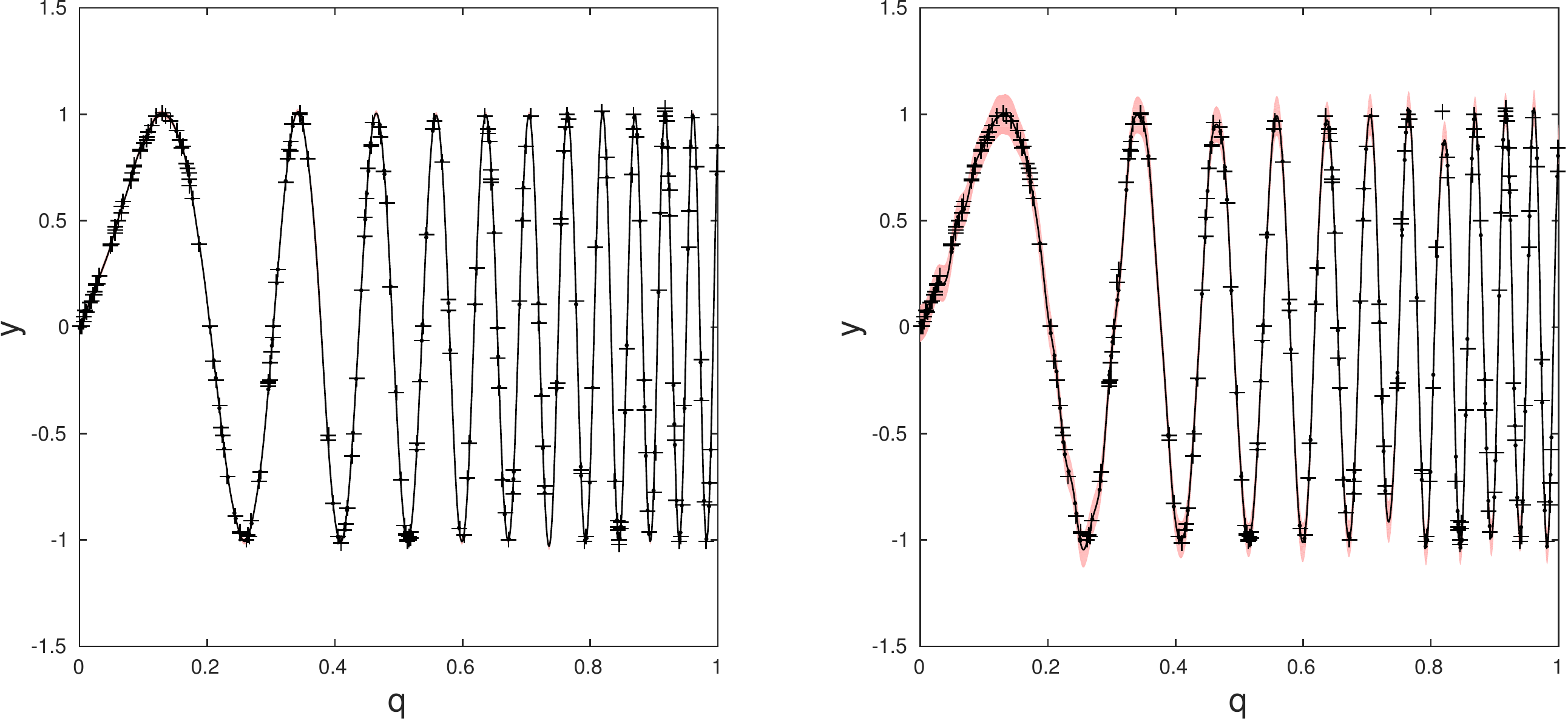}
\caption{Comparison of performance of the standard GP (left) and MGP with $S=1$ (right) for a sine function, $y = \sin(2\pi q(4q + 1)^{1.5})$ with Gaussian noise of standard deviation $\sigma = 0.01$. Crosses represent training points ($N=128$); continuous lines and shaded regions show the predictive mean and the variance.}  \label{sinefig1}
\end{figure}

\begin{figure}[ht]
	\centering
\includegraphics[width=0.95\textwidth]{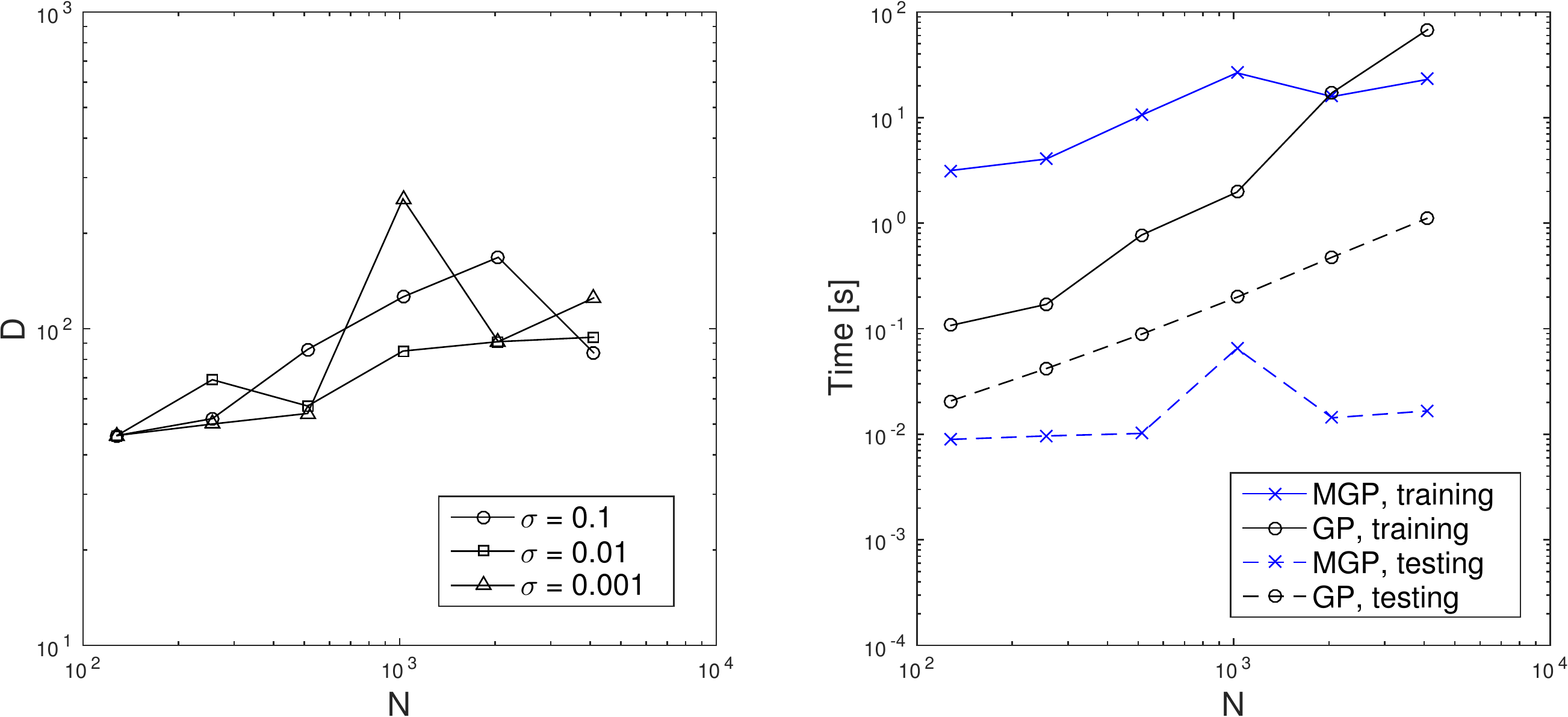}
\caption{Comparison of the performance of the standard GP and MGP for the sine function in Figure \ref{sinefig1}. The
wall-clock time is obtained using MATLAB on a typical desktop PC.}  \label{sinefig2}
\end{figure}

The first numerical example we present is a 1-D step function ($d=1$). We
compare the standard kernel GP with a Gaussian kernel and the present MGP
algorithm restricted to a single scale ($S=1$). The training set of size $N$ was randomly selected from 10000 points distributed
in a uniform grid. Points not used for training were used as test points. The initial data was
generated by adding normally distributed noise with standard deviation $%
\sigma$ to the step function. Optimal hyperparameters were found using the Lagarias algorithm for the Nelder-Mead method (as implemented by MATLAB's
\texttt{fminsearch} function) \cite{lagarias}. For the standard GP, $\sigma$
and $h$ were optimized, while for the MGP algorithm, the optimal $\gamma 
$ was determined in addition to $\sigma$ and $h$. In both cases, the optimal $\sigma$ was close  to the
actual $\sigma $ used for data generation. The ratio between the optimal $h$ for the kernel and the finite-dimensional
(``weight-space'') approaches was found to be approximately 
$1.4$, which is consistent with the difference of $\sqrt{2}$ predicted by
the theory (the distinction between the weight space and function space approaches is provided in Ref.~\cite{rbfgp2}). 

The plots in Figure \ref{stepfig1} show that there is no substantial
difference in the mean and the variance computed using both methods, while
the sparse algorithm required only $D=38$ functions compared to the $N=128$
required for the standard method. Figure~\ref{stepfig1b} shows the relationship between the extended feature space (D) with respect to the size of the training set (N).
 Ideally, this should be a linear dependence, since the step function is scale-independent. Due to random noise and the possibility that the optimization may
converge to local minima of the objective function, however, the relationship is not exactly linear. 
Since $D$ is nevertheless several times smaller than $N$, the wall-clock time for the present
algorithm is shorter than that for the standard GP in both testing and training, per optimizer iteration.
The total training time for the present algorithm can be larger than that of the 
standard algorithm due to the overhead associated with the larger number of hyperparameters and the resultant increase in the number of optimization iterations required. 
However, we observed this only for relatively small values of $N$. For larger $N$, such as $N=4096$, the present
algorithm was approximately 5, 10, and 20 times faster than the standard
algorithm for $\sigma =0.1$, $0.01$, and $0.001$, respectively.

Figure \ref{stepfig2} illustrates a case where the training points are distributed non-uniformly. Such situations
frequently appear in practical problems, where  regions of high functional gradients are sampled with higher density to provide a good representation of the function. For example, adaptive meshes to capture phenomena such as shockwaves and boundary layers in fluid flow fall in this category. 
In the case illustrated here, the training points were
distributed with exponential density near the jump ($q=0.5\pm h_{t}\ln z$, $%
\exp \left( -0.5/h_{t}\right) \leqslant z\leqslant 1$, where $z$ are
distributed uniformly at the nodes of a regular grid. $h_{t}=0.1$ was used). For MGP, the number of scales was $S=6$
and the other hyperparameters were optimized using the same routine as before. Due to multiple extrema in the objective function, it is rather difficult to optimize the number of \ scales, $S$. In practice, one should start from several initial guesses or fix some parameters such as $S$. We used several values of $%
S$ and observed almost no differences for $5\leq S \leq 10$,
while results for $S=1$ and $S=2$ were substantially different from the cases where $S>2$.

It is seen that the MGP provides a much better fit of the step
function in this case than the standard method. This is achieved due to its
broad spectrum of scales. In the present example, we obtained the following
optimal parameters for scales distributed as a geometric progression, $%
h_{s}=h_{1}\beta ^{s-1}$: $h_{\max }=h_{1}=0.1233$, $h_{\min }=h_{6}=0.0032$%
, and $\beta =0.4806$. Other optimal parameters were $\gamma =0.258$ and $\sigma
=0.127$. For the standard GP, the optimal scale was $h=0.0333$. Figure \ref{stepfig2}
shows that with only a single intermediate scale, it is impossible to
approximate the function between training points with a large spacing, whereas MGP provides a much better approximation.
Moreover, since $h_{\min }<h$, we also have a better approximation of the jump, i.e. of
small-scale behavior. This is clearly visible in the figure; the jump for
the standard GP is stretched over about 10 intervals between sampling
points, while the jump for MGP only extends over 3 intervals.
Note that in the present example, we obtained $D=N=101$, so the
wall-clock time for testing is not faster than the standard GP. However, this case
illustrates that multiple scales can provide good results for substantially
non-uniform distributions where one scale description is not sufficient. 

As final analytical example, we explore a sine wave with varying frequency, depicted in Figure \ref{sinefig1}. As before, $N$ training points are randomly chosen from a uniform distribution, and $M=10000-N$ test points are used. For MGP, $S=1$ was used. Compared to the previous examples, this is an intermediate case in terms of scale dependence. One noteworthy result from this dataset is that $D$ is almost constant with respect to $N$, as seen in Figure \ref{sinefig2}. This shows that when the function is relatively smooth, the optimization process is not limited to producing a linear relationship between $D$ and $N$. Another result is that the output variance of the multiscale method is visibly higher than that of the standard method. For the previous cases, the variances have been either been close to equal, or the standard method would produce the higher variance. This could be due to the fact that $h$ and $D$ are inherently related for the current method, whereas $h$ is unrestricted for the standard GP. Since $D<N$, the multiscale $h$ for $S=1$ is typically greater than the standard method's $h$. According to Eqs. (\ref{p11}), this would result in greater variance. 

\subsection{Data from Turbulent Combustion}
Combustion in the presence of turbulent flow involves an enormous disparity in time and length scales, meaning that direct numerical simulations (DNS) are not possible
in practical problems. Large eddy simulations (LES), in which the effect of scales smaller than the mesh resolution ({\em i.e.} subgrid scales) 
are modeled, is often a pragmatic choice. A key difficulty in LES of combustion is the modeling of the subgrid scale fluxes 
in the species transport equations~\cite{kant1,kant2}. These terms
arise as a result of the low-pass filtering - represented by the operator $(\bar{\cdot})$ - of the governing equations, and are of the form 
\begin{equation}
f_k = \overline{\rho u_k C}-\frac{\overline{\rho u_k}\overline{\rho C}}{\bar{\rho}}
\end{equation}
where $\rho, u,C$ represent density, velocity and species mass fraction, respectively. 
Subgrid-scale closures based on concepts from non-reacting flows, such as the equation below,\footnote{$C_s, Sc$ are typically constants, and $S_{ij} = \frac{1}{2}\left[ 
\frac{\partial u_i}{\partial x_j} + \frac{\partial u_j}{\partial x_i} \right]$ is the strain-rate tensor. The superscript $(\tilde{.})$ denotes Favre-filtering
and is defined by $\tilde{q}=\frac{\overline{\rho q}}{\bar{q}}$. $\Delta$ is the filter size.}
\begin{equation}
f_{k} = -\frac{\overline{\rho} C_s^2 \Delta^2 \sqrt{2 \tilde{S}_{ij} \tilde{S}_{ij}}}{Sc}\frac{\partial \tilde{c}}{\partial x_k}
\end{equation} 
are found to be inadequate for turbulent combustion. Modeling of the scalar fluxes thus continues to be an active area of research, and many analytical models are being evaluated by the community. Reference ~\cite{chakraborthy} provides a concise summary of such developments in the area of premixed turbulent flames.


In this work, we intend to apply GP and MGP to identify the model relationships from data. The simulation data is obtained from Ref.~\cite{raman}, in which
DNS of a propagating turbulent flame is performed. The configuration involves a premixed hydrogen-air flame propagating in the 
x-direction. The flame forms a thin front between the burnt and unburnt gases.
Specifically, we attempt to learn the normalized flux in the x-direction, $F_1$, as a function of seven variables:
\begin{equation}
F_1 = \frac{f_1}{\overline{\rho} \Delta^2} = f(\nabla \tilde{u},\nabla \tilde{c},\tilde{S})
\end{equation}

A total of 3 million training points were generated from the dataset by performing low-pass filtering.  Some care was taken in choosing training points. Since the flame is thin relative to the size of the domain, the majority of the data points were found to lie outside the region where $F_1$ is nonzero. To mitigate this disadvantageous distribution, 80 percent of the training points were randomly chosen from the data with $f_1 > 0.05$, and 20 percent were chosen from data with $f_1 \leq 0.05$, i.e. outside the flame. About 2500 training points were used in total. For testing, a single $x-y$ plane of around 6500 points was set aside. 

The predictive results on this plane are shown in Table \ref{comberr}, and Figure \ref{combscatter} shows the ML output versus the true DNS values. From these plots, it is seen that MGP has achieved a ten-fold increase in evaluation speed for a corresponding two percent increase in error. In the scatter plot, MGP appears to perform better than GP for low values of $F_1$ and worse for high values, but the overall difference is small. Figure \ref{combcontour} is a side-by-side comparison of contours of $F_1$ from DNS and from GP and MGP. It is especially evident in the contour plot that GP and MGP are both able to capture features of the flame, whereas analytical models described in Ref.~\cite{chakraborthy} were not as accurate\footnote{These results are not shown}.
 Figure \ref{combslice} plots $F_1$ along several locations perpendicular ($x$) and parallel ($y$) to the flame in the test plane. 

\begin{table}
\caption{Numerical results for learning $F_1$ in the flame. Error is defined as $||\mb{y} - f(\mb{Q}_*)||/||\mb{y}||$, where $\mb{y}$ is a vector of the exact outputs $F_1$, and $f(\mb{Q}_*)$ is a vector of the GP or MGP outputs at the test points $\mb{Q}_*$. Time is test time in seconds, obtained in MATLAB.}
\label{comberr}
\centering
  \begin{tabular}{| l | c | c | c | c |}
    \hline
    & \multicolumn{2}{c|}{GP} & \multicolumn{2}{c|}{MGP} \\ \hline
     & Error & Time & Error & Time \\ \hline
    $F_1$ & 0.1439 & $0.94$ & 0.1566 & $9.9\times10^{-2}$ \\ 
    \hline
  \end{tabular}
\end{table}

\begin{figure}[ht]
	\centering
	\pic{0.6}{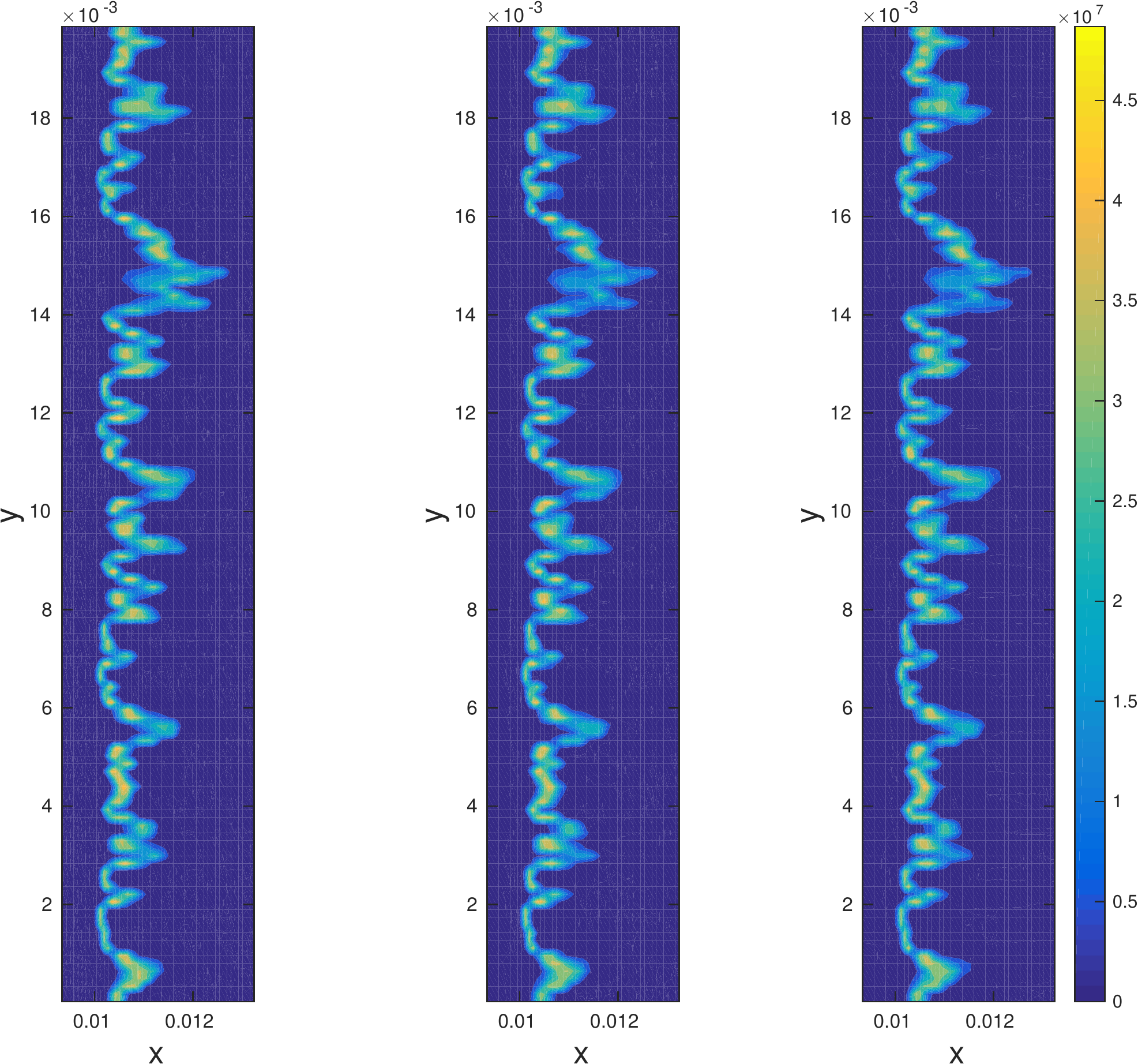}
		\caption{Contours of $F_1$ on the test $x-y$ plane. Left: DNS values (exact). Center: GP output. Right: MGP output.}  \label{combcontour}
\end{figure}

\begin{figure}[ht]
	\centering
	\pic{0.95}{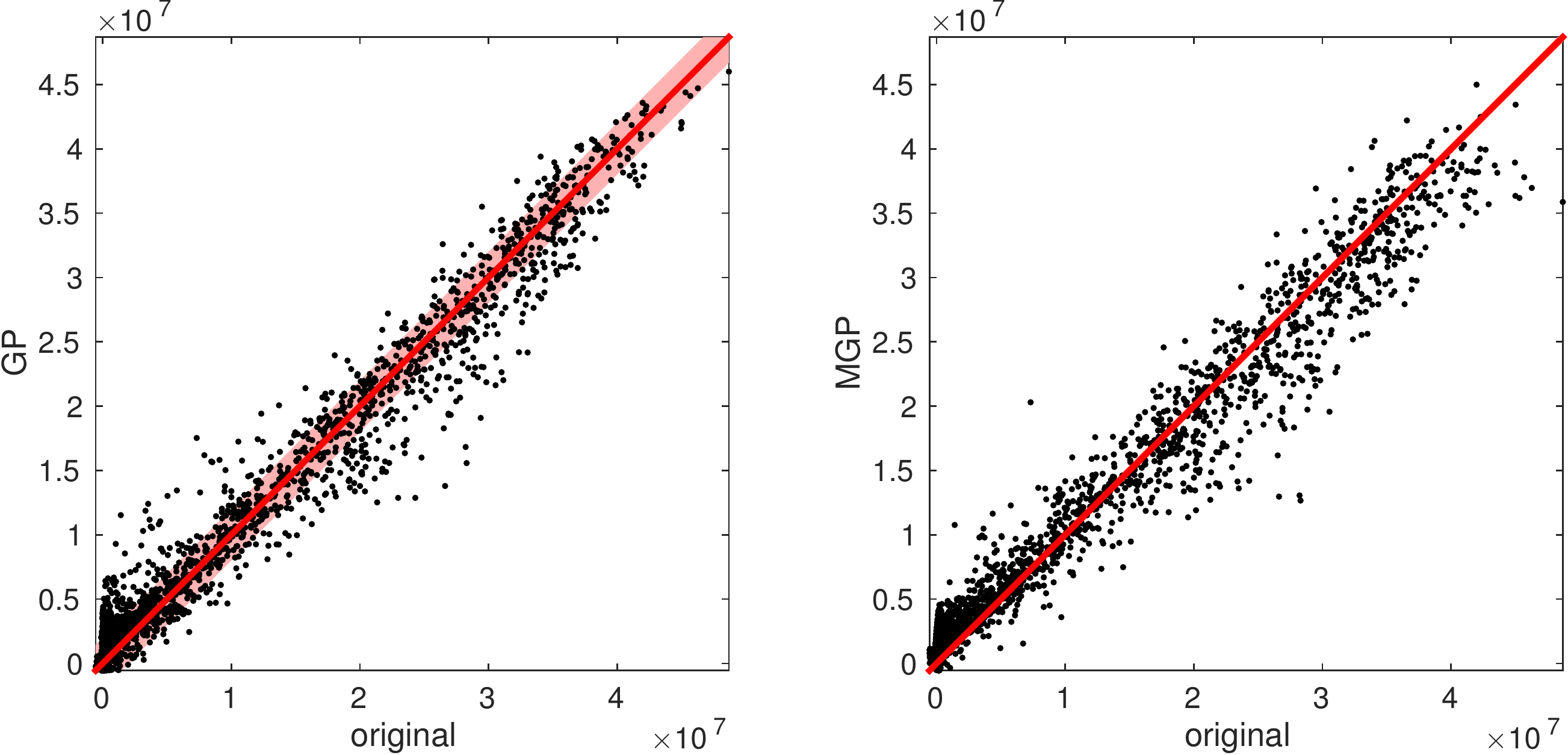}
		\caption{Original test output versus learned result for $F_1$. Red regions are $\pm\sigma$ from the diagonal, i.e. the optimized estimate of the input noise.}  \label{combscatter}
\end{figure}

\begin{figure}[ht]
\centering
	    \subfigure[$y=2\times10^{-5}$]{\includegraphics[width=80mm]{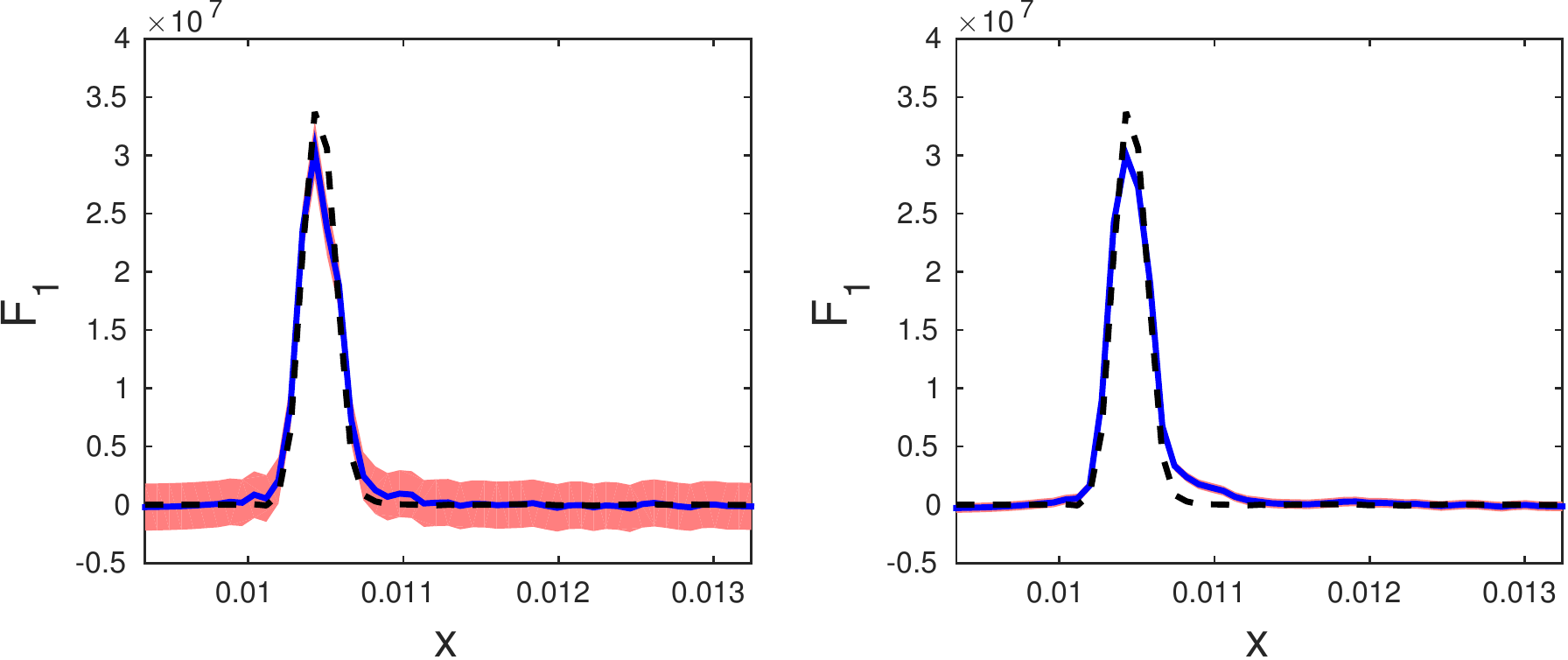}}
	    \subfigure[$y=8\times10^{-4}$]{\includegraphics[width=80mm]{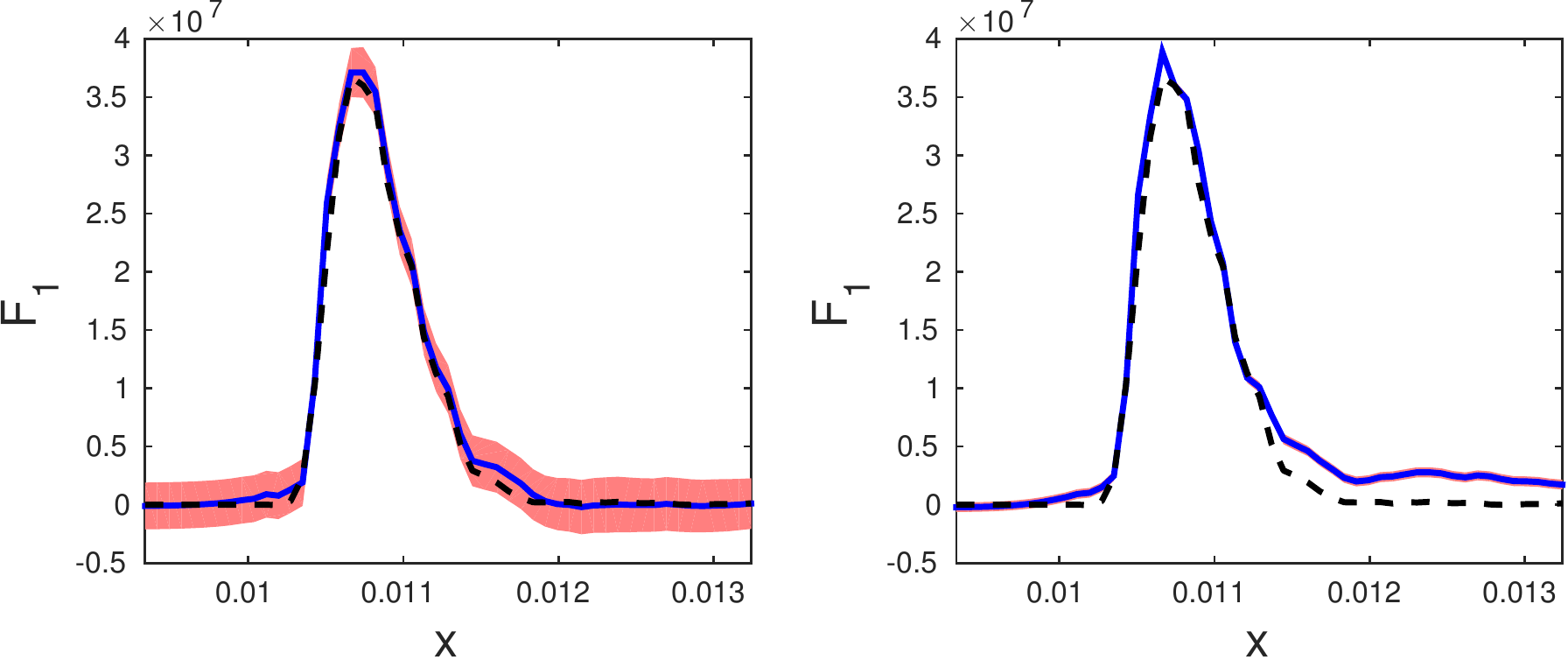}}
	    \subfigure[$y=5\times10^{-3}$]{\includegraphics[width=80mm]{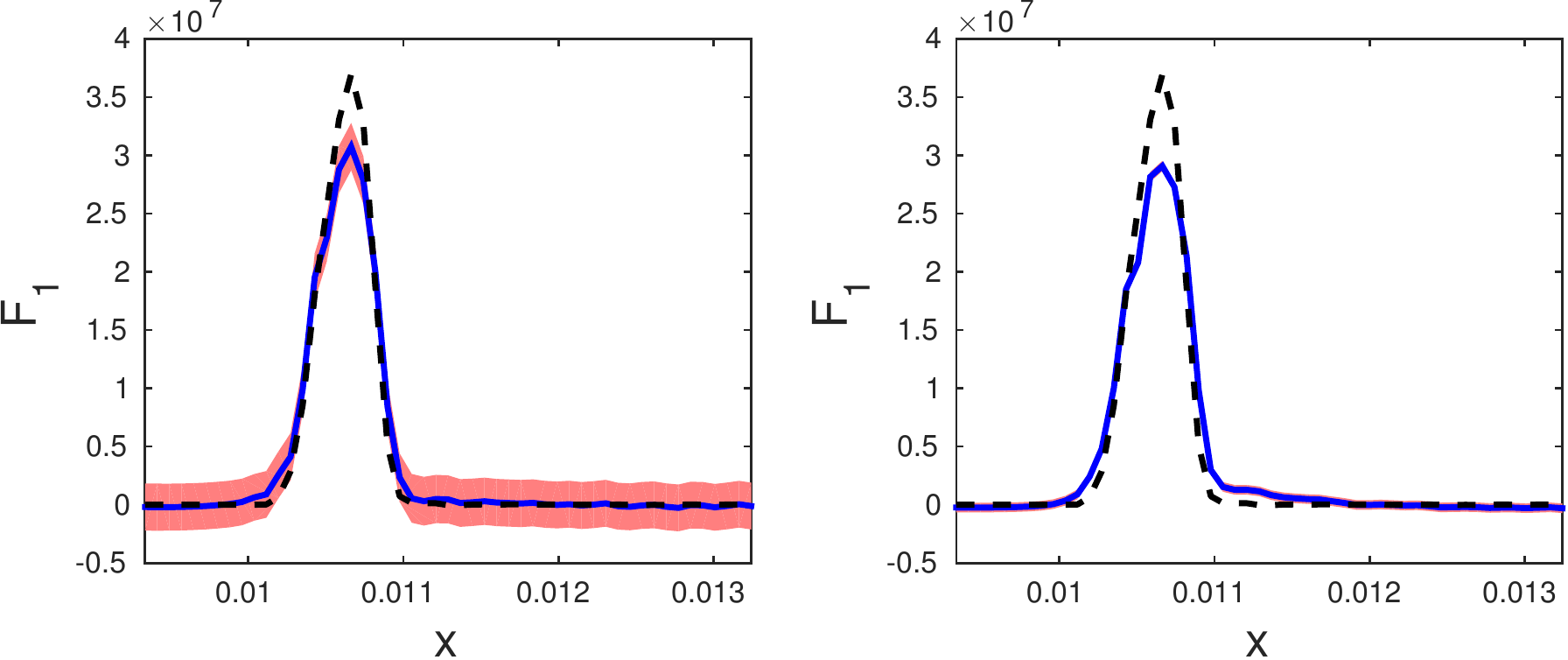}}
	    \subfigure[$y=1\times10^{-2}$]{\includegraphics[width=80mm]{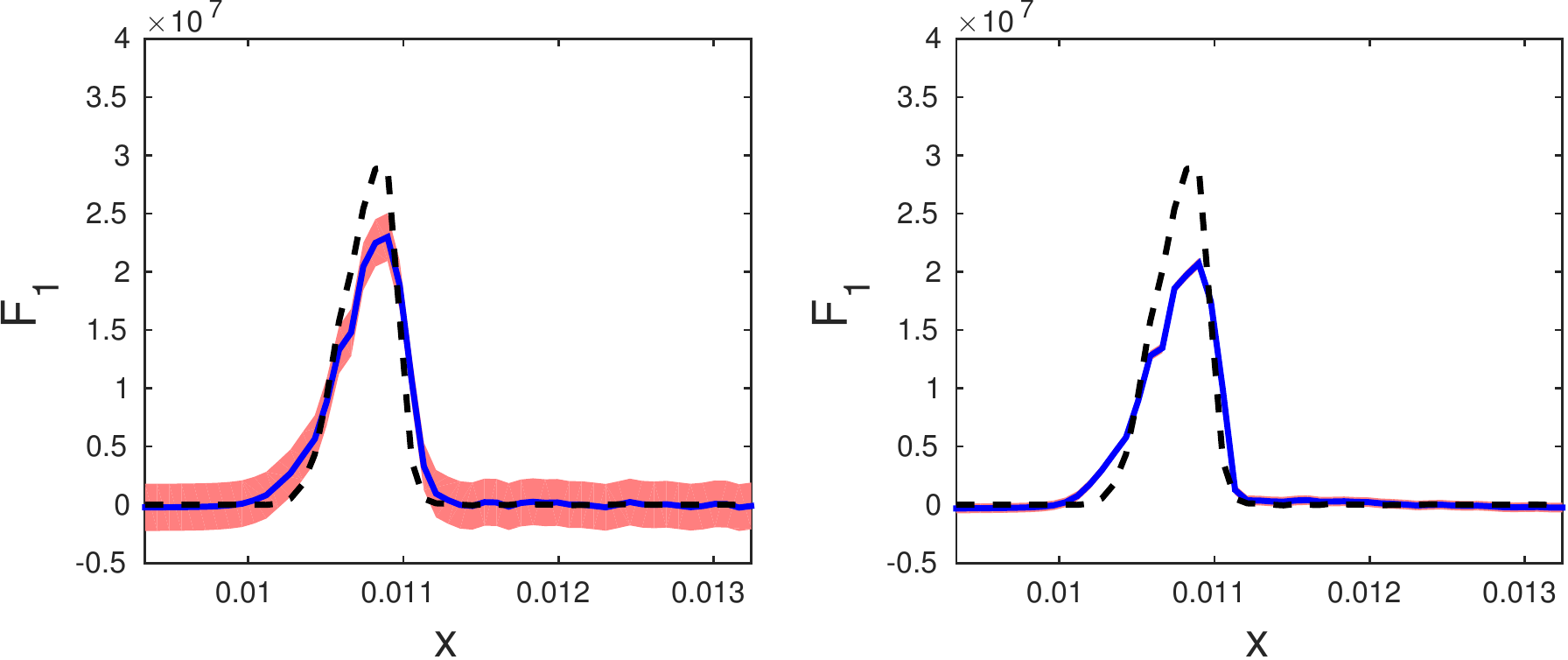}}
		\caption{$F_1$ as a function of $x$ for four different constant values of $y$ on the test $x-y$ plane. Dashed black lines are from DNS. GP results on the left, MGP results on the right. Red regions are one standard deviation for the output.}  \label{combslice}
\end{figure}

\section*{Appendix}
Almost all interpolation problems are intrinsically ill-posed. Assume that just one point $\mb{q}_k$ is removed from the initial set of training points $\mb{Q}_{\mr{train}}$, and an interpolant is constructed based on the $N_{\mr{train}}-1$ values of $y$ at the rest of the points. 
\[f(\mb{q}_k) = \sum_{n=1}^{N_{\mr{train}}-1}\gamma_n\phi(\mb{q}_k,\mb{q}_n)\]
If the ML function is good enough (in terms of smoothness and bounds of its derivatives), the interpolation should produce $f(\mb{q}_k)$ that is close to $y_k$. Hence, two sets of coefficients $\gamma_i$, first from fitting over the complete set, $\gamma_i^{(1)}$ , and second from fitting over the reduced set, $\gamma_i^{(2)}$, should both provide similar values. We note that the length of the set $\gamma_i^{(2)}$ can be made the same as $\gamma_i^{(1)}$ by putting the coefficient corresponding to the missing training point, $\gamma_k^{(2)}$, to 0. As any of the points can play the role of the removed point $\mb{q}_k$, this shows that $\gamma_k^{(1)}$ can be large, while its counterpart $\gamma_k^{(2)} = 0$. In other words, small
variations in point selection result in large variations of $\gamma$, and there exist sets of
very different $\gamma_i$ that provide $\epsilon$-approximate fitting, i.e.
\[\left|y_k - \sum_{n=1}^{N}\gamma_n\phi(\mb{q}_k,\mb{q}_n)\right| < \epsilon, \qquad k = 1,...,N\]

Typical regularization, also referred to as Tikhonov regularization, seeks to minimise the functional
\[F = ||\mb{\Phi\gamma} - \mb{y}|| + \lambda||\mb{\gamma}||, \qquad \lambda > 0\]
If $||\cdot||$ is taken as the 2-norm, the fitting problem becomes
\[(\mb{\Phi} + \lambda\mb{I})\mb{\gamma} = \mb{y}\]
Hence, $\lambda$ can be seen as a regularisation parameter in GP, where the weights $\gamma$ are those in Eq. \ref{generalmatrix}. The concept of regularization comes from kernel ridge regression, which is the non-Bayesian perspective of Gaussian processes. There, the regularization factor $\lambda$ is not explicitly linked to the noise of the input data. 

\section*{Acknowledgments}
This work was performed under the NASA LEARN grant titled ``A Framework for Turbulence Modeling Using Big Data.'' The authors thank Dr. Shivaji Medida, Prof. Juan Alonso, and Mr. Brendan Tracey for fruitful discussions.

\bibliography{refs}
\bibliographystyle{plain}

\end{document}